\documentclass[10pt,twocolumn,letterpaper]{article}

\usepackage[final, algorithms]{wacv}
\usepackage{times}
\usepackage{epsfig}
\usepackage{graphicx}
\usepackage{amsmath}
\usepackage{amssymb}

\usepackage{graphicx}
\usepackage{amsmath}
\usepackage{amssymb}
\usepackage{booktabs}
\usepackage{xspace}
\usepackage{pifont}

\usepackage[pagebackref,breaklinks,colorlinks]{hyperref}
\usepackage[capitalize]{cleveref}
\crefname{section}{Sec.}{Secs.}
\Crefname{section}{Section}{Sections}
\Crefname{table}{Table}{Tables}
\crefname{table}{Tab.}{Tabs.}
\newcommand{\pparagraph}[1]{\medskip\noindent\textbf{#1}}
\usepackage[table]{xcolor}
\usepackage{highlight}

\newcommand{\tp}{\textsc{tp}\xspace}
\newcommand{\tn}{\textsc{tn}\xspace}
\newcommand{\fp}{\textsc{fp}\xspace}
\newcommand{\fn}{\textsc{fn}\xspace}
\newcommand{\pr}{\textsc{pr}\xspace}
\newcommand{\re}{\textsc{re}\xspace}
\newcommand{\fpr}{\textsc{fpr}\xspace}
\newcommand{\pwc}{\textsc{pwc}\xspace}
\newcommand{\fe}{\textsc{f$_1$}\xspace}
\newcommand{\siou}{\textsc{sIoU}\xspace}

\newcommand{\cpos}[1]{\gradientcelld{#1}{-40}{0}{40}{red}{white}{green}{70}}
\newcommand{\cneg}[1]{\gradientcelld{#1}{-40}{0}{40}{green}{white}{red}{70}}
\newcommand{\probP}{\text{I\kern-0.15em P}}

\newcommand{\cross}{\ding{55}}%
\renewcommand{\checkmark}{\ding{51}}%
\crefname{section}{Sec.}{Secs.}
\Crefname{section}{Section}{Sections}
\Crefname{table}{Table}{Tables}
\crefname{table}{Tab.}{Tabs.}

\def\confName{WACV}
\def\confYear{2024}

\begin{document}

\title{Reducing False Alarms in Video Surveillance by Deep Feature Statistical Modeling}

\author{Xavier Bou\and Aitor Artola\and Thibaud Ehret\and Gabriele Facciolo\and Jean-Michel Morel\and Rafael Grompone von Gioi\\
Université Paris-Saclay, CNRS, ENS Paris-Saclay, Centre Borelli\\
91190, Gif-sur-Yvette, France\\
{\tt\small xavier.bou\_hernandez@ens-paris-saclay.fr}
}
\maketitle

\begin{abstract}
   Detecting relevant changes is a fundamental problem of video surveillance. Because of the high variability of data and the difficulty of properly annotating changes, unsupervised methods dominate the field. Arguably one of the most critical issues to make them practical is to reduce their false alarm rate. In this work, we develop a method-agnostic weakly supervised a-contrario validation process, based on high dimensional statistical modeling of deep features, to reduce the number of false alarms of \emph{any} change detection algorithm. We also raise the insufficiency of the conventionally used pixel-wise evaluation, as it fails to precisely capture the performance needs of most real applications. For this reason, we complement pixel-wise metrics with object-wise metrics and evaluate the impact of our approach at both pixel and object levels, on six methods and several sequences from different datasets. Experimental results reveal that the proposed a-contrario validation is able to largely reduce the number of false alarms at both pixel and object levels.
\end{abstract}

\section{Introduction}
\label{sec:intro}

Video change detection 
is a fundamental problem in computer vision and the first step of many applications. While it is an easy task for humans in many contexts, it turns out to be very difficult to automate due to the wide range of possible scenarios.

In the domains of security and surveillance, 
change detection can be used for 
spotting 
temporal anomalies such as suspicious individuals or stolen objects~\cite{suspicious_app, suspicious_app_2}. In urban scenarios, it can be exploited to analyze common activities such as monitoring illegal parking of vehicles~\cite{surv_app_2_parking}. Change detection may also serve climate and humanitarian causes. Satellite image time series can be used to monitor urban development~\cite{ground_vis_a_contrario} of specific regions and the variability of gas concentrations in the atmosphere across time~\cite{methane_plumes, methane_plumes_2}.

\begin{figure}[t]
    \centering
    \includegraphics[width=1\linewidth,trim={8.2cm 5.5cm 9.2cm 5cm},clip]{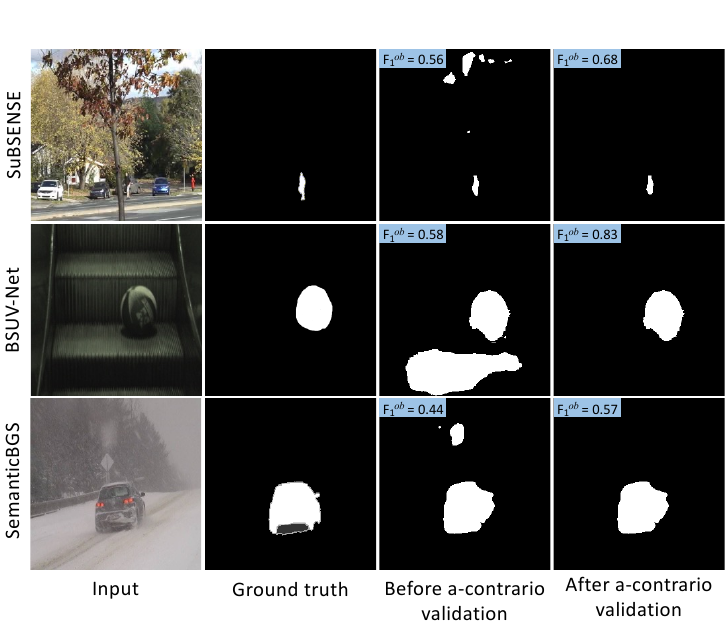} 
    \caption{Sample results of the proposed a-contrario validation process for three different background subtraction algorithms. As shown, the proposed a-contrario validation is able to remove false positives of large and small sizes, while keeping true detections. The object-wise F-1 score of the corresponding sequence is highlighted in each case.}
    \label{fig:teaser_fig}
\end{figure}

Traditional change detection methods work by learning a statistical model of a scene under normal conditions. This so-called background model is based on past samples~\cite{vibe, subsense}. When a new frame is provided, it is compared to the background reference model, which can lead to a detection and/or to an update of this background model. Traditional pixel-wise methods are convenient because only a short local training is required, and the computational complexity remains low. Nevertheless, these techniques are limited by the locality of the features, such as RGB pixels, which makes them prone to false alarms. To overcome this problem, deep learning has been used to leverage the ability of deep neural networks (DNNs) to learn suitable, high-level descriptors of a scene. Despite having shown improvements over traditional video change detection methods~\cite{Droogenbroeck, Babaee, sakkos}, such approaches are constrained by their supervised nature and experience a substantial decrease of performance when tested on out of distribution data. Recently, some works have focused on exploiting the semantic information provided by DNNs in an unsupervised manner~\cite{semanticbgs, cd_unsup_remote_sensing}. These methods are more robust than classical approaches and do not require labeled examples. However, they are still sensitive to false positives in complex environments such as dynamic backgrounds or adverse weather conditions.

Decreasing the number of false positives in unsupervised methods is a high priority goal. Indeed, a significant number of false alarms may saturate a detection system or require human intervention, which is expensive and time consuming. Change detection methods are conventionally evaluated on pixel-wise metrics, regardless of the spatial organization of faulty pixels: multiple small false detections are counted on par with a single false detection with equivalent area. As a result, pixel-wise scores may not realistically represent the performance of target applications. The number of false alarms is better evaluated at the object level than at the pixel level, because the cost of a false alarm is generally independent of its size. Hence, we shall favor object-wise performance metrics, where by object we understand a connected component.

In this work, we present a weakly supervised a-contrario validation process, based on high-dimensional modeling of deep features, to largely reduce the number of false positives at both the pixel and object levels. The contributions of our work are as follows:
\begin{itemize}
\itemsep0em 
\item We propose a method agnostic, weakly supervised a-contrario validation process that can significantly reduce false alarms in video change detection. To the best of our knowledge, this is the first work to use the a-contrario framework at the DNN feature level for a detection problem.
\item We evaluate our work on six methods at both pixel and object levels. Furthermore, we test them on a set of sequences from different datasets, namely CDNet~\cite{2012cdnet, 2014cdnet}, LASIESTA~\cite{lasiesta} and J. Zhong and S. Sclaroff~\cite{iccv03_paper}.
\item Our results show a considerable increase in object-wise performance metrics, while also improving or maintaining the pixel-wise results. Figure \ref{fig:teaser_fig} illustrates this improvement.
\end{itemize}

\section{Related works}
\label{sec_prev_work}
The detection of temporal anomalies in a video sequence or an image time series is known as change detection. It is difficult to find a definition of temporal anomalies that suits all cases. The methods in the literature look for changes with respect to previously observed examples that are semantically meaningful for the desired downstream task. Change detection algorithms in the literature can be categorized into traditional and neural network-based methods.

\pparagraph{Traditional change detection}
Traditional change detection methods use statistical computer vision techniques to model the background of a scene and update it online \cite{SOBRAL20144}. These approaches commonly follow a three-step workflow consisting in 1) building a background model of the scene, 2) comparing the new observed frames to the background model, and 3) updating the model accordingly. Background modeling consists in building a faithful probabilistic representation of the past, which is used as a reference for further observed examples. The seminal example of this approach is the adaptive Gaussian Mixture Model (GMM), first introduced by Stauffer and Grimson~\cite{Stauffer_et_Grimson_vol2}, which modeled each pixel with a mixture of $K$ Gaussian functions. Several modifications of their method have been proposed~\cite{zivkovic, gmm_mod_1, gmm_mod_2} to improve performance and efficiency. 
Later methods proposed to model the background using a buffer of past samples, which can alleviate computational complexity. New samples are compared against the stored examples based on a consensus. The popular unsupervised methods ViBE~\cite{vibe} and SuBSENSE~\cite{subsense} use such consensus-based algorithms. During inference, a new unseen frame is compared to the generated background model using an error metric that maps pixels to either background or foreground clusters, producing a binary mask with this two-level information.

\pparagraph{Deep learning-based change detection methods}
More recent works have exploited current deep learning algorithms to replace one or more steps of the traditional flow. Braham and Van Droogenbroeck~\cite{Droogenbroeck} show that 
the complex background modeling task can be simplified by training a CNN with scene-specific examples. An autoencoder-based architecture called FgSegNet is proposed in~\cite{cd_triplet_supervised}, which adapts a VGG-16~\cite{vgg} architecture into a triplet framework, 
processing images at three different scales. Tezcan \textit{et al.}~\cite{bsuv} proposed BSUV-Net, which trains a scene agnostic network so that it can be tested on new, unseen scenes without individually fine-tuning the network. A newer version of their approach, BSUV-Net 2.0~\cite{bsuv2.0}, was later proposed. The ability of DNNs to learn suitable, high-level descriptors of a scene has proved to yield better results than traditional approaches~\cite{Droogenbroeck, Babaee, sakkos}. Nevertheless, supervised methods require large amounts of annotated data, a tedious and time consuming task. Furthermore, the performance of supervised methods often declines on out of domain examples. Consequently, unsupervised methods are often chosen over recent supervised methods~\cite{bgs_review_real_app, BOUWMANS20198}. For this reason, several recent works have focused on leveraging DNN high-level representations without supervision. Braham \textit{et al.}~\cite{semanticbgs} proposed SemanticBGS, where a classic method is complemented with semantic information provided by a pre-trained network. Moreover, a real-time version of the same approach named RT-SemanticBGS~\cite{rtsbgs} was later introduced. G-LBM, introduced by Rezaei \textit{et al.}~\cite{G-LBM}, models the background of a scene with a generative adversarial network (GAN) in the presence of noise and sparse outliers.

Unsupervised DNN-based methods achieve a better performance than traditional approaches. Nonetheless, they 
still fall behind supervised methods in popular benchmarks. Similarly to classic methods, these techniques may detect a substantial number of false alarms, which can critically saturate detection systems.

\begin{figure*}[t]
    \centering
    \includegraphics[width=1\linewidth,trim={2cm 5.7cm 2cm 5.2cm},clip]{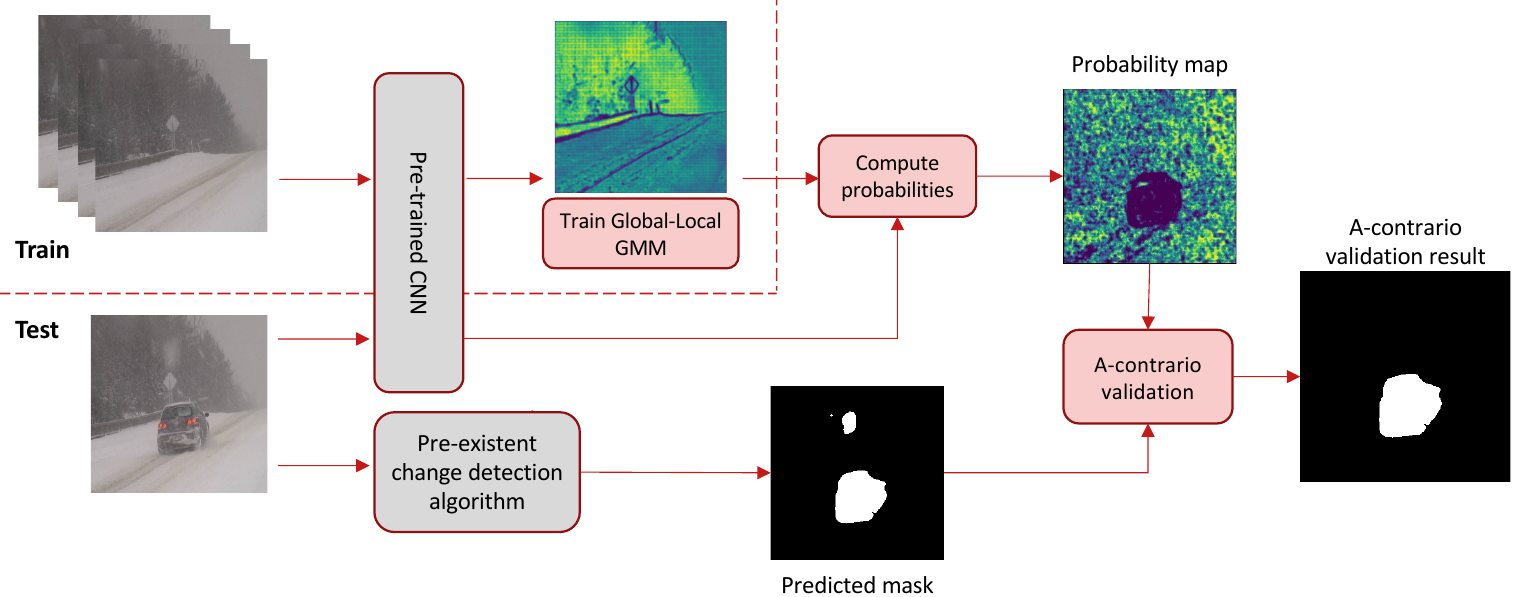}
    \caption{The proposed deep feature a-contrario validation extracts the representations given by a pre-trained network and models them with a global-to-local mixture of Gaussians. For a new, unseen frame, it computes the probability map of observing a temporal anomaly using the trained GMM. Given a change mask predicted by any algorithm in the literature, a validation process based on a-contrario theory addresses all detected regions and removes false alarms.}
    \label{fig:method_diagram}
\end{figure*}

\pparagraph{Statistical modeling for video scene understanding}
Surveillance applications need to distinguish foreground from background elements so that target instances can be further processed. Early methods attempted to statistically model appearance information with parametric models such as 
GMMs~\cite{Stauffer_et_Grimson_vol2, zivkovic,gmm_mod_1, gmm_mod_2}. 
However, background modeling in complex scenarios (e.g. dynamic backgrounds or adverse weather conditions) has proved to be challenging for those approaches. Other works introduce optical flow to understand the motion patterns of a scene. For example, Saleemi \textit{et al.}~\cite{cvpr10_modelling} proposed to model the motion patterns with a mixture of Gaussians using optical flow. Similarly, Ghahremannezhad \textit{et al.}~\cite{modeling_flow2} proposed a method for real-time foreground segmentation modeling optical flow with a GMM. Some recent works have propsoed to model the feature space of deep neural networks for image anomaly detection. PaDiM, proposed by Defard \textit{et al.}~\cite{padim}, is a framework that models patches of a DNN feature map with a Gaussian model for anomaly detection and localization. Artola \textit{et al.}~\cite{artola2023glad} generalized this attempt with GLAD, a method that learns a robust GMM globally, and then localizes the learned Gaussians with a spatial weight map. Modeling spatial features in a high-dimensional space has shown promising results at recognizing complex patterns.

\pparagraph{A-contrario detection theory}
The a-contrario detection theory is a mathematical formulation of the non-accidentalness principle, which states that an observed structure is meaningful only when the relation between its parts is too regular to be the result of an accidental arrangement of independent parts~\cite{witkin_1, lowe}. 
The a-contrario methodology~\cite{meaningful_alignments, meaningful_alignments_2}
allows one to control the number of false alarms by considering an observed 
structure only when the expectation of its occurrences is small in a stochastic background model.
The 
Number of False Alarms (NFA) of an event $e$ observed up to a precision $z(e)$ in the background model $\mathcal{H}_0$ is defined by
\begin{equation} \label{eq_a_contrario}
\mathrm{NFA}(e) = N_T \cdot \probP[Z_{\mathcal{H}_0}(e) \geq z(e)],
\end{equation}
where $\probP[Z_{\mathcal{H}_0}(e) \geq z(e)]$ is the probability of obtaining a precision $Z_{\mathcal{H}_0}(e)$ 
better or equal to the observed one $z(e)$ in the background model $\mathcal{H}_0$. The term $N_T$ corresponds to the number of tests, following the statistical multiple hypothesis testing framework~\cite{hyp_testing}. A small NFA indicates that the event $e$ is unlikely to be randomly observed in the background model $\mathcal{H}_0$. Hence, the lower the NFA the more meaningful the event. 
A value $\epsilon$ is specified and candidates with $\mathrm{NFA} < \epsilon$ are accepted as valid detections. It can be shown~\cite{meaningful_alignments} that in these conditions $\epsilon$ is an upper-bound to the expected number of false detections under $\mathcal{H}_0$.

A-contrario methods have been previously proposed 
to address computer vision problems. Lisani and Ramis developed a method in~\cite{a_contrario_faces} that applied an a-contrario methodology on a normal distribution for the detection of faces in images. In surveillance, a-contrario methods have been used mainly in the remote sensing field, \cite{ground_vis_a_contrario, a_contrario_super_pixel}, where the temporal difference between images is large, and no tracking of temporal objects is feasible. Grompone \etal~\cite{ground_vis_a_contrario} proposed an a-contrario method based on a uniform distribution and a greedy algorithm to compute candidate regions, detecting visible ground areas in satellite imagery.

\section{Pixel and object-wise evaluation}
\label{evaluation_cd}
Change detection algorithms tend to be evaluated on pixel-wise metrics. Whilst this approach allows one to assess how well methods classify pixels into foreground or background clusters, it often fails to represent the performance needs of real applications. Detection systems today tend to consider detections as sets of connected components instead of independent pixels. Then, they process each detection separately for further analysis. An algorithm with high pixel-wise evaluation scores might still predict a considerable amount of false alarms at the object level, which can lead to bottlenecks in the system. Focusing on the performance at the object level will provide a more accurate account of the usability of methods for surveillance applications. Hence, reducing false positives at the object level is one of the most important elements to minimize in order to increase speed and avoid system saturation.

Consequently, we consider both pixel-wise and object-wise evaluation metrics to analyze the performance of our work and existing algorithms. We emphasize our evaluation on the reduction of false alarms and the accuracy of the detections. Let \tn, \tp, \fn, \fp be the usual pixel-wise number of true negatives, true positives, false negatives and false positives, respectively. Our experiments consider the following pixel-wise metrics:
\begin{itemize}
\itemsep0em
  \item Precision: $\pr^{pi} = \tp / (\tp + \fp)$
  \item Recall: $\re^{pi} = \tp / (\tp+\fn)$
  \item False Positive Rate:  $\fpr^{pi} = \fp/(\fp+\tn)$
  \item Percentage of Wrong Classifications:\\ $\pwc^{pi} = 100 \times (\fn+\fp) / (\tp+\fn+\fp+\tn)$
  \item F-measure: $\fe^{pi}=2 (\pr^{pi} \times \re^{pi}) / (\pr^{pi} + \re^{pi})$.
\end{itemize}
We also evaluate the results at the object level, where $\tp^{ob}$, $\fn^{ob}$ and $\fp^{ob}$ now correspond to true positives, false negatives and false positives for sets of connected components. To define those, we take the approach of Chan \etal~\cite{segment_me} and use a variation of the traditional intersection over union (IoU) first introduced by Rottman \etal~\cite{siou}. Unlike the conventional IoU, which penalizes cases where a ground truth region is fragmented into multiple predictions by assigning each prediction a moderate IoU score, the adapted metric, named sIoU, does not penalize predictions of a segment when the remaining ground truth is sufficiently covered by other predicted segments. 


More formally, let $\mathcal{K}$ be the set of anomalous components in the ground truth, and $\hat{\mathcal{K}}$ the set of anomalous components predicted by a change detection algorithm. hen, the sIoU metric consists in a mapping $sIoU: \mathcal{K} \rightarrow [0, 1]$ defined for $k \in \mathcal{K}$ by

\begin{equation} 
\label{eq_siou}
\begin{split}
sIoU(k) := \frac{| k\cap \hat{\mathcal{K}}(k) |}{| (k\cup \hat{\mathcal{K}}(k) )\backslash \mathcal{A}(k) |} \\
\text{with} \quad \hat{\mathcal{K}}(k) = \bigcup_{\hat{k}\in\hat{\mathcal{K}}, \hat{k}\cap k \neq \varnothing} \hat{k},
\end{split}
\end{equation}
where $\mathcal{A}(k) = \{z\in k^{\prime} : k^{\prime} \in \mathcal{K} \backslash \{k\}\}$. The introduction of $\mathcal{A}(k)$ excludes all pixels from the union if and only if they correctly intersect with another ground-truth component. Hence, given a threshold $\tau \in [0, 1)$, we define a target $k\in\mathcal{K}$ as $\tp^{ob}$ if $sIoU > \tau$, and as $\fn^{ob}$ otherwise. Then, $\fp^{ob}$ is computed as the positive predictive value (PPV) for $\hat{k}\in \hat{\mathcal{K}}$, defined as:
\begin{equation} \label{eq_ppv}
PPV(\hat{k}) := \frac{|\hat{k} \cap \mathcal{K}(\hat{k})|}{|\hat{k}|}.
\end{equation}
Thus, $\hat{k}\in \hat{\mathcal{K}}$ is $\fp^{ob}$ if $PPV(\hat{k}) \leq \tau$. Lastly, the sIoU-based F-measure is computed as follows:
\begin{itemize}
\itemsep0em
  \item F-measure: $\fe^{sIoU}(\tau)=\frac{\tp^{ob}(\tau)}{\tp^{ob}(\tau)+\fn^{ob}(\tau)+\fp^{ob}(\tau)}$
\end{itemize}
We follow the approach of Chan \etal and average the results for different thresholds $\tau = \{0.25, 0.5, 0.75\}$.

\section{Our approach}\label{sec_method}

We propose to supplement any change detection method from the literature with a final a-contrario validation step applied on connected components. Such validation is based on a statistical model of the DNN representations of the scene and requires no annotation. Given a set of input frames of a sequence, we obtain its feature representations at stages 1 and 2 of a pre-trained ResNet-50~\cite{resnet} architecture. We use a backbone pre-trained on ImageNet~\cite{imagenet} with self-supervision using the VicReg method~\cite{vicreg}. Figure~\ref{fig:method_diagram} shows a high-level diagram of the proposed approach, and the next subsections describe its different parts in detail.

\subsection{Background feature modeling} 
We model the extracted deep representations with a mixture of Gaussians to assess the likelihood of an image patch to be a part of the background. For that we extend the GLAD~\cite{artola2023glad} framework to the background modeling of videos. This requires no dense annotations but just a selection of training frames with none or few anomalies present, thus we coin our approach as weakly supervised. A mixture is first learned globally, i.e. without taking into consideration the spatial location of the data points. This yields a first Gaussian mixture model $\theta = (\phi_i, \mu_i, \Sigma_i)_{i \in \{1,\dots, K\}}$,
where $\mu_i$ and $\Sigma_i$ are the mean and variance components, while $\phi_i$ are the mixture weights.
Then, a local model is derived by assigning position-dependent weights for each Gaussian, so that an image position is represented by a local mixture of the most relevant Gaussian distributions.
This gives a localized model that depends on the pixel position $(x,y)$ such that 
$\theta(x,y) = (\phi_i(x,y), \mu_i, \Sigma_i)_{i \in \{1,\dots, K\}}$, where $\mu_i$ and $\Sigma_i$ do not depend on the position $(x,y)$.
This global to local approach enables one to exploit information from other similar pixels, to build a good representation of each observed pixel. 
The probability of observing $\textbf{p}$ at position $(x,y)$ is 
\begin{equation}\label{eq:gmm}
    \mathbb{P}\big(\textbf{p}~|~\theta(x,y)\big) = \sum_{i=1}^K \phi_i(x,y) \mathbb{P}(\textbf{p}~|~\mu_i, \Sigma_i),
\end{equation}
where only the weights $\phi_i(x,y)$ depend on the position.  Then, the $p\mbox{-}\mathrm{value}$ of a given pixel $\textbf{p}$ at position $(x,y)$ is
\begin{equation}\label{eq:pvalue}
    p\mbox{-}\mathrm{value}\big(\textbf{p}~|~\theta(x,y)\big) =
    \hspace{-12mm}
    \int\limits_{\hspace{12mm}\mathcal{D}(\textbf{p}|\theta(x,y))}
    \hspace{-10mm}
    \mathbb{P}\big(\textbf{q}~|~\theta(x,y)\big) \, d\textbf{q},
\end{equation}
where $\mathcal{D}(\textbf{p}|\theta(x,y)) =\big\{\textbf{q}~|~\mathbb{P}(\textbf{q}|\theta(x,y)) \leq \mathbb{P}(\textbf{p}|\theta(x,y))\big\}$.
This quantity cannot be easily computed, but an upper-bound can be derived. We can invert the sum of the GMM and the integral of the $p$-value to get $K$ Gaussian integrals. However the set $\mathcal{D}$ cannot be computed, so we introduce  $\mathcal{D}_i(\textbf{p}|\theta(x,y)) =\big\{\textbf{q}~|~ \phi_i(x,y) \mathbb{P}(\textbf{q}~|~\mu_i, \Sigma_i) \leq \mathbb{P}(\textbf{p}|\theta(x,y))  \big\}$ which contains it ($\mathcal{D}\subseteq\mathcal{D}_i$). So we find ourselves with the upper bound
\begin{equation}\label{eq:pvalue-upper-bound}
    p\mbox{-}\mathrm{value}\big(\textbf{p}~|~\theta(x,y)\big) \leq  
    \sum_{i=1}^K
    \hspace{-13mm}
    \int\limits_{\hspace{13mm}
    \mathcal{D}_i(\textbf{p}|\theta(x,y))}
    \hspace{-10mm}
    \mathbb{P}(\textbf{q}~|~\mu_i, \Sigma_i) \, d\textbf{q}.
\end{equation}
These integrals are equivalent to the $\chi^2$ survival function; in the case where the features are of even dimension they are equal to a finite sum that can be computed exactly (see the supplementary material).

\subsection{A-contrario validation}

We propose an a-contrario method designed to control the number of false alarms on any change detection method. A change candidate will only be meaningful when the expectation of all its independent parts is low. In our case, we define the background model $\mathcal{H}_0$ as the local Gaussian mixture $\theta$ learned during training. Thus, we define the number of false alarms $\mathrm{NFA}(R)$ 
over a region $R$, as:
\begin{equation} \label{eq_7}
\mathrm{NFA}(R) = N_T \cdot \probP[Z_{\theta}(R) \geq z(R)],
\end{equation}
%
where $z(R)$ measures how anomalous the observed values in region $R$ are. The corresponding random variable $Z_{\theta}$ is a random vector with the same dimension as the number of pixels of $R$. Then, assuming pixel independence (see below), we propose to measure anomaly relative to a Gaussian mixture $\theta$ by defining the probability term of \eqref{eq_7} as:
%
%
%
\begin{equation} \label{eq_8}
\probP[Z_{\theta}(R) \geq z(R)] = \prod_{\textbf{p}\in R} p\mbox{-}\mathrm{value}\big(\textbf{p}~|~\theta(x,y)\big),
\end{equation}
where $p\mbox{-}\mathrm{value}\big(\textbf{p}~|~\theta(x,y)\big)$, given by Equation~\eqref{eq:pvalue}, is evaluated on each pixel $\textbf{p}$ of the region $R$.

We need to define the number of tests $N_T$ to complete the NFA formulation~\eqref{eq_7}.
$N_T$ is related to the total number of candidate regions that can, in theory, be considered for evaluation. Inspired by the approach in~\cite{ground_vis_a_contrario}, we will consider regions of any shape formed by 
4-connected pixels. Regions of pixels with 4-connectivity are known as polyominoes~\cite{polyminoes, hyp_testing}. 
The exact number $b_n$ of different polyomino configurations of a given size $n$ is not known in general; however, a good estimate~\cite{jensen2000statistics} is given by
$b_n \approx \alpha \frac{\beta^n}{n}$
where $\alpha \approx 0.317$ and $\beta \approx 4.06$. Additionally, we need to consider that any pixel in the image can be the center of a region and that a region can be of size from 1 to $XY$, where $X$ and $Y$ are the width and height of the image, respectively. Thus, we can define the number of tests $N_T$ as
\begin{equation} \label{eq_10}
N_T = X^2Y^2 \alpha \frac{\beta^n}{n},
\end{equation}
where $n=|R|$ is the size of the region $R$.
Notice that this is not exact, as Equation~\eqref{eq_10} allows for some potential polyominoes extending outside of the image boundaries, but it is an approximation of the same magnitude. 

Deep feature representations are computed via spatial convolutions. Hence, assuming pixel independence over an entire region is inaccurate. To compensate for this, we introduce a correcting exponent $\frac 1{c_f}$ where $c_f$ corresponds to the area of the receptive field of a given stage, i.e. $c_f=35$ for stage 1 and $c_f=91$ for stage 2. 
We then compute a geometric mean of the probability term on the receptive field.
All in all, Equation~\eqref{eq_7} can be expressed as
\begin{equation} \label{eq_11}
\mathrm{NFA}(R) = X^2Y^2 \alpha \frac{\beta^n}{n} \left(\prod_{\textbf{p}\in R} 
p\mbox{-}\mathrm{value}\big(\textbf{p}~|~\theta(x,y)\big)\right)^{\frac{1}{c_f}}\hspace{-5mm}
\end{equation}
A region $R$ with $\mathrm{NFA}(R) < \epsilon$ is declared a change detection.
Since the $\mathrm{NFA}$ can take very extreme values, $\log(\mathrm{NFA})$ is often easier to handle.



\section{Experiments}
\label{sec_experiments}
\begin{figure*}[t]
    \centering
    \includegraphics[width=1\linewidth,trim={0.3cm 5.2cm 2.1cm 4.75cm},clip]{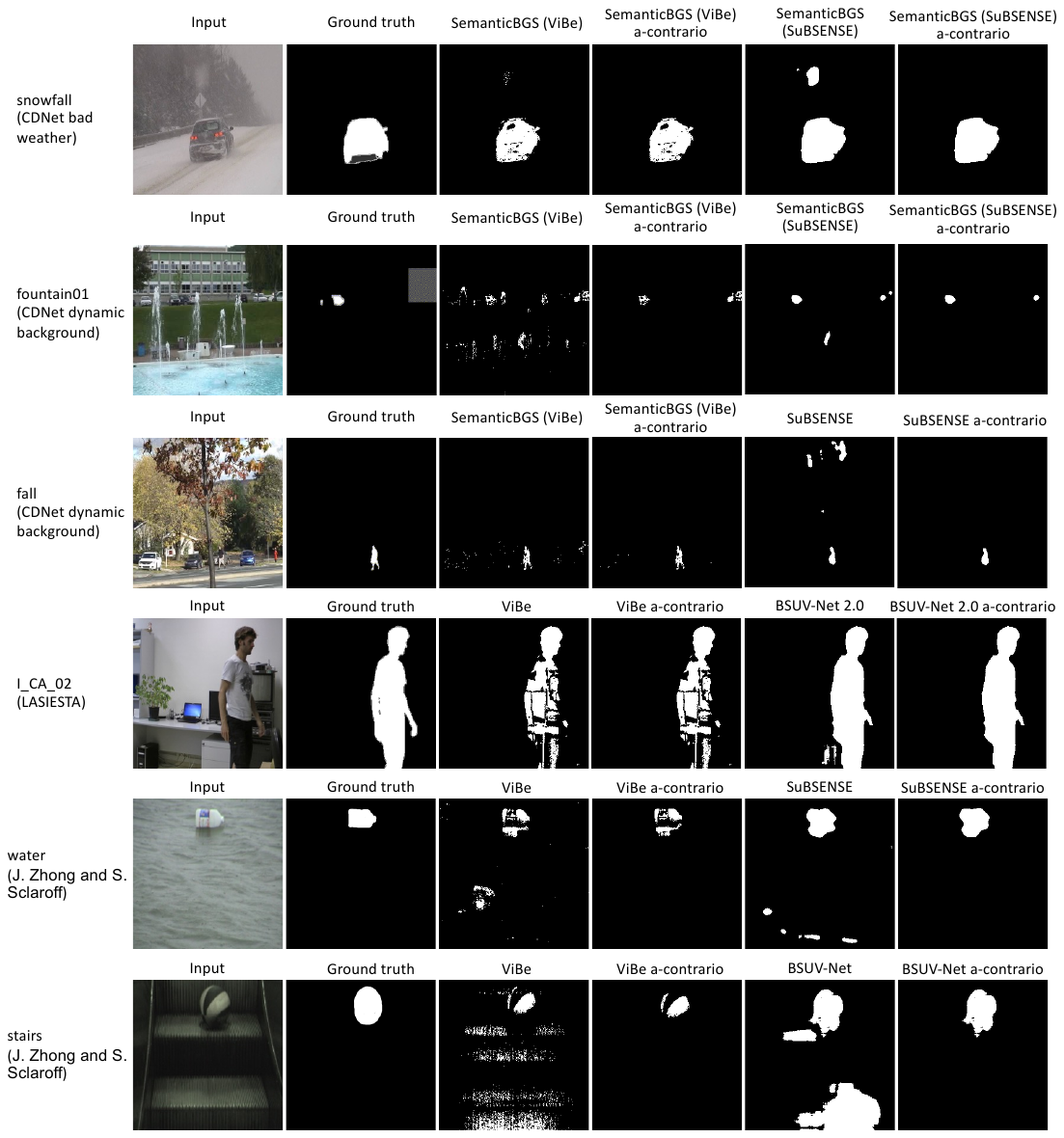}
    \caption{Detection results before and after the a-contrario validation, on sequences taken in different datasets. For each sequence, results are shown for two of the evaluated change detection algorithms.}
    \label{fig:qualitative_results}
\end{figure*}

\pparagraph{Implementation details}
We trained a GMM on each sequence setting $K=1000$ initial Gaussian distributions and we let GLAD remove unnecessary Gaussians, as addressed by its authors as the best option. For training, we selected the first few hundred subsequent frames of each scene without any obvious anomalies. For the sequences where anomalies are continuously present (e.g. cars driving by a road), a temporal median filter is applied to filter them out. For the scenes where first frames contain only a few outliers, we kept them in the training set. The a-contrario validation was applied with a threshold $\epsilon=1$, and we computed the geometric mean of the NFA scores of each stage. 
For simplicity, the sequences and their ground truths were resized to 256$\times$256.

\pparagraph{Results on the CDNet benchmark}
CDNet~\cite{2012cdnet, 2014cdnet} is a benchmark for video change detection consisting of 53 videos divided into 11 categories. Each category corresponds to a specific challenge such as \textit{shadow} or \textit{lowFramerate}. It provides pixel-wise annotations for all frames except the first few hundred, which can be used for initialization and we use to train our models. We compute the proposed approach on all categories of the dataset except for categories \textit{cameraJitter} and \textit{PTZ}, since we assume the camera is static.
We evaluated the impact of our approach on a range of methods for the pixel and object-wise metrics introduced in Section~\ref{evaluation_cd}. The selected methods were ViBe~\cite{vibe}, SuBSENSE~\cite{subsense}, SemanticBGS~\cite{semanticbgs}, BSUV-Net~\cite{bsuv}, BSUVFPM-Net~\cite{bsuv} and BSUV-Net 2.0~\cite{bsuv2.0}. 
The classic ViBe and SuBSENSE are still popular and remain on the state of the art. SemanticBGS is the best performing unsupervised method in CDNet, and the BSUV family of methods are supervised, scene agnostic approaches. Table~\ref{table:cdnet_results} shows the relative change percentage when applying the proposed a-contrario validation compared to the methods alone on all the CDNet dataset (except \textit{cameraJitter} and \textit{PTZ}). Furthermore, we selected two categories that are substantially prone to false alarms: \textit{dynamic background} and \textit{bad weather}, and show their results separately on Table~\ref{table:cdnet_dyb_results} and Table~\ref{table:cdnet_badw_results}.
For the tables containing the exact metric scores, refer to the supplementary material. Formally, we denote by $S_b$ and $S_{a}$ the metric scores before and after the a-contrario validation, respectively. We then compute the percentage improvement as
\begin{equation} \label{relative_change}
\text{ \textit{relative change percentage} } = \frac{S_{a} - S_b}{S_b}\times100.
\end{equation}
As observed, our approach is able to largely improve the object-wise metrics in all methods, while maintaining or improving their pixel-wise counterparts. Notice that in some cases, such as for ViBe, the percentage improvement is surprisingly large. This occurs as a consequence of removing all FPs that correspond to isolated pixels, as the method commonly yields significant segmentation noise in complex scenes. Figure~\ref{fig:qualitative_results} shows examples of the achieved qualitative results for different sequences and methods. Lastly, the proposed method proved to be robust to anomalies in the training data, yielding weak clusters and low probabilities for such occurrences, as per the cases where a few anomalies were present in the training set.

\renewcommand{\arraystretch}{1.2}

\begin{table}[t]
\centering

\resizebox{\columnwidth}{!}{
\begin{tabular}{p{3cm}ccccc}
\hline  
Method                        &  \multicolumn{3}{c}{pixel-wise}   & \multicolumn{2}{c}{object-wise}  \\

                            \cmidrule[0.4pt](lr{0.125em}){2-4}%
                            \cmidrule[0.4pt](lr{0.125em}){5-6}%
                            
                            & $\fpr^{pi}\downarrow$   & $\pwc^{pi}\downarrow$      & $\fe^{pi}\uparrow$     & $\siou\uparrow$    &  $\fe^{sIoU}\uparrow$    \\

                            \cmidrule[0.4pt](lr{0.125em}){2-2}%
                            \cmidrule[0.4pt](lr{0.125em}){3-3}%
                            \cmidrule[0.4pt](lr{0.125em}){4-4}%
                            \cmidrule[0.4pt](lr{0.125em}){5-5}%
                            \cmidrule[0.4pt](lr{0.125em}){6-6}%
                           
ViBe~\cite{vibe}                        & \cneg{-64.45}\%	& \cneg{-22.81}\%	& \cpos{13.40}\%	& \cpos{204.00}\%	& \cpos{36.12}\%           \\
SuBSENSE~\cite{subsense}                    & \cneg{-20.36}\%	& \cneg{-3.06}\%	& \cpos{-1.20}\%	& \cpos{14.96}\%	& \cpos{8.69}\%        \\
SemanticBGS (with ViBe)~\cite{semanticbgs}     & \cneg{-41.10}\%	& \cneg{-6.70}\%	& \cpos{3.93}\%	& \cpos{119.07}\%	& \cpos{95.50}\%       \\
SemanticBGS (with SuBSENSE)~\cite{semanticbgs} & \cneg{-19.05}\%	& \cneg{-0.15}\%	& \cpos{-1.73}\%	& \cpos{31.13}\%	& \cpos{19.78}\%   \\
BSUV-Net 2.0~\cite{bsuv2.0}                & \cneg{-11.61}\%	& \cneg{3.88}\%	& \cpos{-1.74}\%	& \cpos{24.86}\%	& \cpos{9.45}\%            \\
BSUV-Net FPM~\cite{bsuv}                & \cneg{0.02}\%	& \cneg{2.02}\%	& \cpos{-0.78}\%	& \cpos{9.75}\%	& \cpos{3.06}\%                        \\
BSUV-Net~\cite{bsuv}                    & \cneg{-3.13}\%	& \cneg{1.21}\%	& \cpos{-1.19}\%	& \cpos{16.60}\%	& \cpos{7.13}\%                \\ \hline  
\end{tabular}
}
\caption{Results on the CDNet benchmark, which includes all categories with the exception of \textit{cameraJitter} and \textit{PTZ}, since we assume the camera is static. Both pixel-wise and obejct-wise metrics are provided.}
\label{table:cdnet_results}

\end{table}

\begin{table}[t]
\centering

\resizebox{\columnwidth}{!}{
\begin{tabular}{p{3cm}ccccc}
\hline  
Method                        &  \multicolumn{3}{c}{pixel-wise}   & \multicolumn{2}{c}{object-wise}  \\

                            \cmidrule[0.4pt](lr{0.125em}){2-4}%
                            \cmidrule[0.4pt](lr{0.125em}){5-6}%
                            
                            & $\fpr^{pi}\downarrow$   & $\pwc^{pi}\downarrow$      & $\fe^{pi}\uparrow$     & $\siou\uparrow$    &  $\fe^{sIoU}\uparrow$    \\

                            \cmidrule[0.4pt](lr{0.125em}){2-2}%
                            \cmidrule[0.4pt](lr{0.125em}){3-3}%
                            \cmidrule[0.4pt](lr{0.125em}){4-4}%
                            \cmidrule[0.4pt](lr{0.125em}){5-5}%
                            \cmidrule[0.4pt](lr{0.125em}){6-6}%
                           
ViBe~\cite{vibe}                        & \cneg{-88.97}\%     & \cneg{-93.07}\%     & \cpos{158.96}\%     & \cpos{5159.21}\%     & \cpos{5350.29}\%      \\
SuBSENSE~\cite{subsense}                    & \cneg{-27.25}\%     & \cneg{-3.00}\%      & \cpos{1.21}\%       & \cpos{37.99}\%     & \cpos{28.28}\%  \\
SemanticBGS (with ViBe)~\cite{semanticbgs}     & \cneg{-82.42}\%     & \cneg{-77.27}\%     & \cpos{31.40}\%      & \cpos{1583.91}\%     & \cpos{1472.55}\%   \\
SemanticBGS (with SuBSENSE)~\cite{semanticbgs} & \cneg{-1.60}\%      & \cneg{-0.03}\%      & \cpos{0.01}\%       & \cpos{30.09}\%     & \cpos{21.92}\%  \\
BSUV-Net 2.0~\cite{bsuv2.0}                & \cneg{-6.36}\%      & \cneg{-0.55}\%      & \cpos{-0.02}\%   & \cpos{49.20}\%     & \cpos{32.77}\%  \\
BSUV-Net FPM~\cite{bsuv}                & \cneg{27.41}\%      & \cneg{18.07}\%      & \cpos{-0.63}\%      & \cpos{8.46}\%     & \cpos{2.76}\%      \\
BSUV-Net~\cite{bsuv}                    & \cneg{-3.29}\%      & \cneg{0.11}\%       & \cpos{-0.22}\%      & \cpos{30.26}\%     & \cpos{16.44}\%        \\ \hline  
\end{tabular}
}
\caption{Results on the \textit{dynamic background} category of the CDNet benchmark. Both pixel-wise and obejct-wise metrics are provided.}
\label{table:cdnet_dyb_results}

\end{table}

\begin{table}[t]
\centering

\resizebox{\columnwidth}{!}{
\begin{tabular}{p{3cm}cccccc}
\hline  
Method                        &  \multicolumn{3}{c}{pixel-wise}   & \multicolumn{2}{c}{object-wise}  \\

                            \cmidrule[0.4pt](lr{0.125em}){2-4}%
                            \cmidrule[0.4pt](lr{0.125em}){5-6}%
                            
                            & $\fpr^{pi}\downarrow$   & $\pwc^{pi}\downarrow$      & $\fe^{pi}\uparrow$     & $\siou\uparrow$    &  $\fe^{sIoU}\uparrow$    \\

                            \cmidrule[0.4pt](lr{0.125em}){2-2}%
                            \cmidrule[0.4pt](lr{0.125em}){3-3}%
                            \cmidrule[0.4pt](lr{0.125em}){4-4}%
                            \cmidrule[0.4pt](lr{0.125em}){5-5}%
                            \cmidrule[0.4pt](lr{0.125em}){6-6}%
                           
ViBe~\cite{vibe}                                             & \cneg{-85.43}\%     & \cneg{-13.38}\%     & \cpos{21.98}\%      & \cpos{301.73}\%    & \cpos{331.69}\%    \\
SuBSENSE~\cite{subsense}                    & \cneg{-11.02}\%     & \cneg{-1.04}\%      & \cpos{0.38}\%       & \cpos{22.20}\%     & \cpos{15.65}\%      \\
SemanticBGS (with ViBe)~\cite{semanticbgs}     & \cneg{-50.35}\%     & \cneg{-0.58}\%      & \cpos{-0.48}\%      & \cpos{234.33}\%      & \cpos{227.09}\%    \\
SemanticBGS (with SuBSENSE)~\cite{semanticbgs} & \cneg{-9.68}\%      & \cneg{-0.57}\%      & \cpos{0.09}\%       & \cpos{61.75}\%     & \cpos{57.21}\%   \\
BSUV-Net 2.0~\cite{bsuv2.0}                & \cneg{-1.36}\%      & \cneg{0.51}\%       & \cpos{-0.17}\%   & \cpos{33.77}\%     & \cpos{13.00}\%          \\
BSUV-Net FPM~\cite{bsuv}                & \cneg{-0.72}\%      & \cneg{0.14}\%       & \cpos{-0.10}\%      & \cpos{12.21}\%     & \cpos{5.50}\%           \\
BSUV-Net~\cite{bsuv}                    & \cneg{-3.24}\%      & \cneg{0.38}\%       & \cpos{-0.17}\%      & \cpos{28.06}\%     & \cpos{12.79}\%          \\ \hline  
\end{tabular}
}
\caption{Results on the \textit{bad weather} category of the CDNet benchmark. Both pixel-wise and obejct-wise metrics are provided.  }
\label{table:cdnet_badw_results}
\end{table}

\pparagraph{Results on LASIESTA sequences}
The LASIESTA (Labeled and Annotated Sequences for Integral Evaluation of SegmenTation Algorithms) dataset~\cite{lasiesta} is a fully annotated benchmark for change detection and foreground segmentation proposed by Cuevas \textit{et al.} It is composed of different indoor and outdoor scenes organized in categories, such as \textit{camouflage}, \textit{shadows} or \textit{dynamic background}. Consequently, we evaluate all sequences with the exception of the ones with camera motion and simulated motion, as we assume the camera is stable. We train on the first 50 frames, without checking whether they contain desired detections. The impact of our approach is evaluated in Table~\ref{table:lasiesta_results}. In this case, SemanticGBS is not considered because the required semantic segmentation pre-computed maps are only provided for CDNet sequences. We can observe that most methods alone already reach good pixel-wise scores and that the a-contrario validation does not decrease them. Furthermore, object-wise metrics improve in all cases. This demonstrates the efficacy of our approach in scenarios with both indoor and outdoor scenes.

\begin{table}[t]
\centering

\resizebox{\columnwidth}{!}{
\begin{tabular}{lcccccc}
\hline  
Method                        &  \multicolumn{3}{c}{pixel-wise}   & \multicolumn{2}{c}{object-wise}  \\

                            \cmidrule[0.4pt](lr{0.125em}){2-4}%
                            \cmidrule[0.4pt](lr{0.125em}){5-6}%
                            
                            & $\fpr^{pi}\downarrow$   & $\pwc^{pi}\downarrow$      & $\fe^{pi}\uparrow$     & $\siou\uparrow$    &  $\fe^{sIoU}\uparrow$    \\

                            \cmidrule[0.4pt](lr{0.125em}){2-2}%
                            \cmidrule[0.4pt](lr{0.125em}){3-3}%
                            \cmidrule[0.4pt](lr{0.125em}){4-4}%
                            \cmidrule[0.4pt](lr{0.125em}){5-5}%
                            \cmidrule[0.4pt](lr{0.125em}){6-6}%
                           
ViBe~\cite{vibe}             & \cneg{-2.07}\% & \cneg{-0.82}\% & \cpos{0.26}\%  & \cpos{58.67}\%  & \cpos{44.75}\%     \\
SuBSENSE~\cite{subsense}     & \cneg{-0.02}\% & \cneg{0.02}\%  & \cpos{-0.02}\% & \cpos{1.00}\%   & \cpos{0.38}\%       \\
BSUV-Net 2.0~\cite{bsuv2.0}  & \cneg{-0.14}\% & \cneg{0.00}\%  & \cpos{-0.01}\%  & \cpos{10.16}\%  & \cpos{3.38}\%       \\
BSUV-Net FPM~\cite{bsuv}     & \cneg{-0.14}\% & \cneg{-0.01}\% & \cpos{0.00}\%  & \cpos{5.30}\%  & \cpos{1.82}\%       \\
BSUV-Net~\cite{bsuv}         & \cneg{-0.16}\% & \cneg{0.02}\%  & \cpos{-0.01}\% & \cpos{7.65}\%  & \cpos{2.79}\%      \\ \hline  
\end{tabular}
}

\caption{Results of the impact of our approach in a set of change detection methods on the selected sequences of LASIESTA dataset. Both pixel-wise and obejct-wise metrics are provided.}
\label{table:lasiesta_results}

\end{table}

\pparagraph{Results on sequences from J. Zhong and S. Sclaroff~\cite{iccv03_paper}}
Lastly, we evaluated our method on two sequences from J. Zhong and S. Sclaroff~\cite{iccv03_paper} containing challenging dynamic backgrounds. 
One of them shows an escalator moving continuously, while the other displays a floating plastic bottle.
In each case, sequences with and without the target objects are provided; we used the latter for training and tested on the former. We evaluated our approach on the same methods as for LASIESTA sequences. The results, reported in Table~\ref{table:iccv03_results}, show a clear improvement of both pixel-wise and object-wise scores in all methods with the exception of BSUV 2.0, where the pixel-wise F-score drops slightly.

\begin{table}[t]
\centering

\resizebox{\columnwidth}{!}{
\begin{tabular}{lcccccc}
\hline  
Method                        &  \multicolumn{3}{c}{pixel-wise}   & \multicolumn{2}{c}{object-wise}  \\

                            \cmidrule[0.4pt](lr{0.125em}){2-4}%
                            \cmidrule[0.4pt](lr{0.125em}){5-6}%
                            
                            & $\fpr^{pi}\downarrow$   & $\pwc^{pi}\downarrow$      & $\fe^{pi}\uparrow$     & $\siou\uparrow$    &  $\fe^{sIoU}\uparrow$    \\

                            \cmidrule[0.4pt](lr{0.125em}){2-2}%
                            \cmidrule[0.4pt](lr{0.125em}){3-3}%
                            \cmidrule[0.4pt](lr{0.125em}){4-4}%
                            \cmidrule[0.4pt](lr{0.125em}){5-5}%
                            \cmidrule[0.4pt](lr{0.125em}){6-6}%
                           
ViBe~\cite{vibe}                                             & \cneg{-96.44} \%     & \cneg{-83.28} \%     & \cpos{83.28} \%     & \cpos{2312.40} \%     & \cpos{3365.97} \%    \\
SuBSENSE~\cite{subsense}     & \cneg{-87.99}\%     & \cneg{-70.06}\%     & \cpos{30.50}\%     & \cpos{379.91}\%      & \cpos{205.67}\%      \\
BSUV-Net 2.0~\cite{bsuv2.0} & \cneg{-54.04}\%     & \cneg{0.32}\%       & \cpos{-4.85}\%     & \cpos{70.46}\%      & \cpos{15.02}\%    \\
BSUV-Net FPM~\cite{bsuv} & \cneg{-88.02}\%     & \cneg{-80.95}\%     & \cpos{39.17}\%     & \cpos{99.97}\%     & \cpos{40.90}\%      \\
BSUV-Net~\cite{bsuv}     & \cneg{-80.93}\%     & \cneg{-77.95}\%     & \cpos{30.04}\%      & \cpos{81.93}\%     & \cpos{30.49}\%       \\ \hline  
\end{tabular}
}
\caption{Results of the impact of our approach in a set of change detection methods on the two sequences from J. Zhong and S. Sclaroff~\cite{iccv03_paper}. Both pixel-wise and obejct-wise metrics are provided.}
\label{table:iccv03_results}
\end{table}

\begin{figure}[t]
    \centering
    \includegraphics[width=1\linewidth]{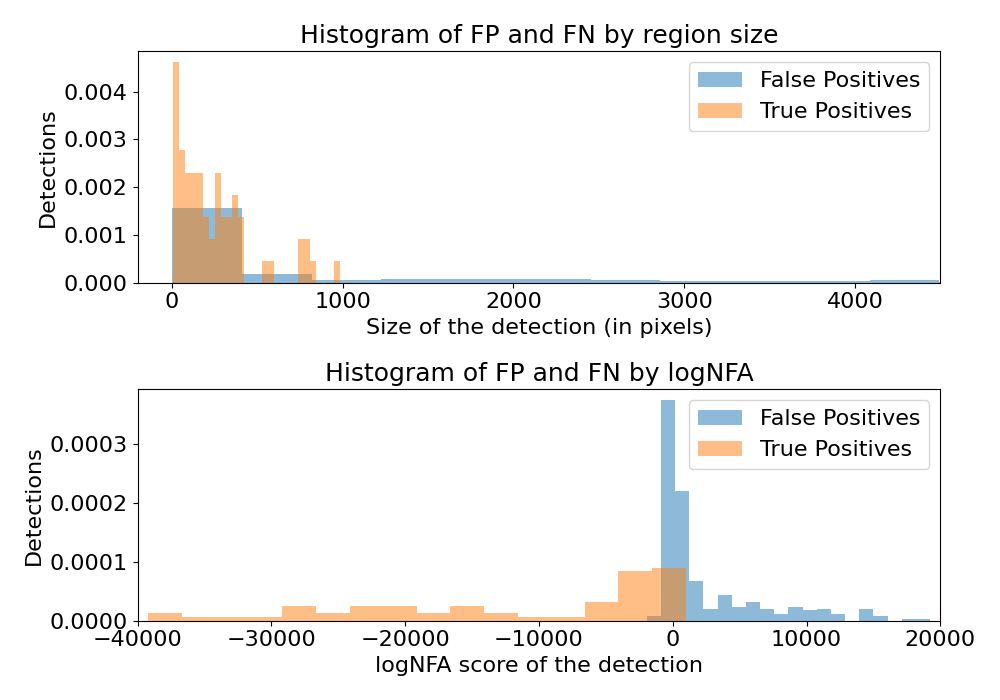}
    \vspace{-2em}
    \caption{Comparison of the histograms of TP and FP computed by the SuBSENSE algorithm for the sequence \textit{escalator}, by size of the detection (top) and by the $\log(\mathrm{NFA})$ score (bottom). As shown, filtering out small detections is not efficient. The $\log(\mathrm{NFA})$ provides a more suitable separation.}
    \label{fig:histograms}
\end{figure}

\pparagraph{Could the obtained drastic reduction of false alarms be achieved by a simpler method than the a-contrario test?} An obvious argument that comes to mind is that eliminating small sized objects might lead to similar performance. It could indeed be argued that most datasets contain large true positives, compared to the size of predicted false positives. Therefore, we checked if simply removing all small detections could lead to significant improvement of the false alarm rate. To study the capability of our approach in discriminating FPs from TPs regardless of the size of the detections, we analyzed their separability in relation to the size and the proposed $\log(\mathrm{NFA})$ score. Doing so also leads to find an optimal value of the threshold $\epsilon$. Figure~\ref{fig:histograms} compares the separability of FPs and TPs based on the region size against the $\log(\mathrm{NFA})$ score, for the results predicted by SuBSENSE for the sequence \textit{escalator}. This particular example shows how FPs and TPs are not separable by region size, but the a-contrario assessment provides a clear separation instead. Additional examples are provided in the supplementary material, for other sequences and methods. The value of $\log(\epsilon)$ is then set to 0 (thus $\epsilon$=1), which provides a reasonable separation of FPs and TPs without discarding a high number of true detections. Moreover, the selection of $\epsilon$=1 holds a strong significance as it corresponds to an expected number of false alarms equal to one.

\section{Conclusion}
\label{sec_discussion}
In this work, we introduced statistical modeling approach of deep features for the reduction of false alarms in video surveillance applications.
In Section \ref{sec_experiments} a series of experiments were conducted to measure the impact of the proposed a-contrario validation on several change detection algorithms. The results indicate that substantial improvement is achieved in object-wise measures in virtually all cases, without decreasing the pixel-wise results. The ability to reduce the number of false alarms by such large margins without hampering detection accuracy is an important step towards real automation of surveillance systems. Moreover, we showed the capability of our approach to discard FPs regardless of their size. To the best of our knowledge, this is the first work that uses statistical modeling in the deep feature space for video surveillance applications.

\pparagraph{Limitations and future work}
While our work successfully decreases the number of false alarms and improves several algorithms on short and mid size sequences, each sequence needs to be trained offline. This compromises the method in long, evolving sequences. Hence, future work will focus on the an online version to adapt to such cases.

{\small
\bibliographystyle{ieee_fullname}
\bibliography{egbib}

\begin{thebibliography}{10}\itemsep=-1pt

\bibitem{artola2023glad}
Aitor Artola, Yannis Kolodziej, Jean-Michel Morel, and Thibaud Ehret.
\newblock Glad: A global-to-local anomaly detector.
\newblock In {\em 2023 IEEE Winter Conference on Applications of Computer
  Vision (WACV)}. IEEE, 2023.

\bibitem{Babaee}
Mohammadreza Babaee, Duc~Tung Dinh, and Gerhard Rigoll.
\newblock A deep convolutional neural network for background subtraction.
\newblock {\em CoRR}, abs/1702.01731, 2017.

\bibitem{vicreg}
Adrien Bardes, Jean Ponce, and Yann LeCun.
\newblock Vicreg: Variance-invariance-covariance regularization for
  self-supervised learning.
\newblock {\em CoRR}, abs/2105.04906, 2021.

\bibitem{vibe}
Olivier Barnich and Marc Van~Droogenbroeck.
\newblock Vibe: A universal background subtraction algorithm for video
  sequences.
\newblock {\em IEEE Transactions on Image Processing}, 20(6):1709--1724, 2011.

\bibitem{BOUWMANS20198}
Thierry Bouwmans, Sajid Javed, Maryam Sultana, and Soon~Ki Jung.
\newblock Deep neural network concepts for background subtraction:a systematic
  review and comparative evaluation.
\newblock {\em Neural Networks}, 117:8--66, 2019.

\bibitem{semanticbgs}
Marc Braham, Sebastien Pierard, and Marc Van~Droogenbroeck.
\newblock Semantic background subtraction.
\newblock In {\em IEEE International Conference on Image Processing (ICIP)},
  pages 4552--4556, Beijing, China, September 2017.

\bibitem{Droogenbroeck}
Marc Braham and Marc Van~Droogenbroeck.
\newblock Deep background subtraction with scene-specific convolutional neural
  networks.
\newblock In {\em 2016 International Conference on Systems, Signals and Image
  Processing (IWSSIP)}, pages 1--4, 2016.

\bibitem{segment_me}
Robin Chan, Krzysztof Lis, Svenja Uhlemeyer, Hermann Blum, Sina Honari, Roland
  Siegwart, Pascal Fua, Mathieu Salzmann, and Matthias Rottmann.
\newblock Segmentmeifyoucan: A benchmark for anomaly segmentation.
\newblock {\em arXiv preprint arXiv:2104.14812}, 2021.

\bibitem{rtsbgs}
Anthony Cioppa, Marc~Van Droogenbroeck, and Marc Braham.
\newblock Real-time semantic background subtraction.
\newblock In {\em 2020 IEEE International Conference on Image Processing
  (ICIP)}, pages 3214--3218, 2020.

\bibitem{lasiesta}
Carlos Cuevas, Eva~María Yáñez, and Narciso García.
\newblock Labeled dataset for integral evaluation of moving object detection
  algorithms: Lasiesta.
\newblock {\em Computer Vision and Image Understanding}, 152:103--117, 2016.

\bibitem{padim}
Thomas Defard, Aleksandr Setkov, Angelique Loesch, and Romaric Audigier.
\newblock Padim: a patch distribution modeling framework for anomaly detection
  and localization.
\newblock In {\em International Conference on Pattern Recognition}, pages
  475--489. Springer, 2021.

\bibitem{meaningful_alignments_2}
Agnès Delsolneux, Lionel Moisan, and Jean-Michel Morel.
\newblock {\em From Gestalt Theory to Image Analysis: A Probabilistic
  Approach}, volume~34.
\newblock Springer Science \& Business Media, 2008.

\bibitem{imagenet}
Jia Deng, Wei Dong, Richard Socher, Li-Jia Li, Kai Li, and Li Fei-Fei.
\newblock Imagenet: A large-scale hierarchical image database.
\newblock In {\em 2009 IEEE Conference on Computer Vision and Pattern
  Recognition}, pages 248--255, 2009.

\bibitem{meaningful_alignments}
Agnès Desolneux, Lionel Moisan, and Jean-Michel Morel.
\newblock Meaningful alignments.
\newblock {\em International Journal of Computer Vision}, 40:7--23, 2000.

\bibitem{bgs_review_real_app}
Belmar Garcia-Garcia, Thierry Bouwmans, and Alberto~Jorge {Rosales Silva}.
\newblock Background subtraction in real applications: Challenges, current
  models and future directions.
\newblock {\em Computer Science Review}, 35:100204, 2020.

\bibitem{modeling_flow2}
Hadi Ghahremannezhad, Hang Shi, and Chengjun Liu.
\newblock Real-time hysteresis foreground detection in video captured by moving
  cameras.
\newblock In {\em 2022 IEEE International Conference on Imaging Systems and
  Techniques (IST)}, pages 1--6, 2022.

\bibitem{polyminoes}
Solomon~W. Golomb.
\newblock {\em Polyominoes: Puzzles, Patterns, Problems, and Packings - Revised
  and Expanded Second Edition}.
\newblock Princeton University Press, 2020.

\bibitem{hyp_testing}
Alexander Gordon, Galina Glazko, Xing Qiu, and Andrei Yakovlev.
\newblock Control of the mean number of false discoveries, bonferroni and
  stability of multiple testing.
\newblock {\em The Annals of Applied Statistics}, 1, 2007.

\bibitem{2012cdnet}
Nil Goyette, Pierre-Marc Jodoin, Fatih Porikli, Janusz Konrad, and Prakash
  Ishwar.
\newblock Changedetection. net: A new change detection benchmark dataset.
\newblock In {\em 2012 IEEE computer society conference on computer vision and
  pattern recognition workshops}, pages 1--8. IEEE, 2012.

\bibitem{ground_vis_a_contrario}
Rafael Grompone, Charles Hessel, Tristan Dagobert, Jean-Michel Morel, and Carlo
  de Franchis.
\newblock Ground visibility in satellite optical time series based on a
  contrario local image matching.
\newblock {\em Image Processing On Line}, 11:212--233, 2021.

\bibitem{gmm_mod_1}
Michael Harville.
\newblock A framework for high-level feedback to adaptive, per-pixel,
  mixture-of-gaussian background models.
\newblock In Anders Heyden, Gunnar Sparr, Mads Nielsen, and Peter Johansen,
  editors, {\em Computer Vision --- ECCV 2002}, pages 543--560, Berlin,
  Heidelberg, 2002. Springer Berlin Heidelberg.

\bibitem{resnet}
Kaiming He, Xiangyu Zhang, Shaoqing Ren, and Jian Sun.
\newblock Deep residual learning for image recognition.
\newblock In {\em 2016 IEEE Conference on Computer Vision and Pattern
  Recognition (CVPR)}, pages 770--778, 2016.

\bibitem{suspicious_app}
Tatsuya Ishikawa and Thi~Thi Zin.
\newblock {\em A Study on Detection of Suspicious Persons for Intelligent
  Monitoring System}, pages 292--301.
\newblock Springer, 2019.

\bibitem{jensen2000statistics}
Iwan Jensen and Anthony~J Guttmann.
\newblock Statistics of lattice animals (polyominoes) and polygons.
\newblock {\em Journal of Physics A: Mathematical and General}, 33(29):L257,
  2000.

\bibitem{cd_triplet_supervised}
Long~Ang Lim and Hacer Keles.
\newblock Foreground segmentation using a triplet convolutional neural network
  for multiscale feature encoding.
\newblock {\em Pattern Recognition Letters}, 112, 2018.

\bibitem{a_contrario_faces}
Jose-Luis Lisani and Silvia Ramis.
\newblock A contrario detection of faces with a short cascade of classifiers.
\newblock {\em Image Processing On Line}, 9:269--290, 2019.

\bibitem{lowe}
David~G. Lowe.
\newblock {\em Perceptual Organization and Visual Recognition}.
\newblock Kluwer Academic Publishers, 1985.

\bibitem{suspicious_app_2}
Khrystyna Lyubymenko, Milan Adamek, and Lukas Kralik.
\newblock Detection of suspicious persons and special software.
\newblock In {\em 2017 12th Iberian Conference on Information Systems and
  Technologies (CISTI)}, pages 1--4, 2017.

\bibitem{surv_app_2_parking}
Matthias Michael, Christian Feist, Florian Schuller, and Marc Tschentscher.
\newblock Fast change detection for camera-based surveillance systems.
\newblock In {\em 2016 IEEE 19th International Conference on Intelligent
  Transportation Systems (ITSC)}, pages 2481--2486, 2016.

\bibitem{gmm_mod_2}
Anurag Mittal and Dan Huttenlocher.
\newblock Scene modeling for wide area surveillance and image synthesis.
\newblock In {\em Proceedings IEEE Conference on Computer Vision and Pattern
  Recognition. CVPR 2000 (Cat. No.PR00662)}, volume~2, pages 160--167 vol.2,
  2000.

\bibitem{cd_unsup_remote_sensing}
Hyeoncheol Noh, Jingi Ju, Minseok Seo, Jongchan Park, and Dong-Geol Choi.
\newblock Unsupervised change detection based on image reconstruction loss.
\newblock In {\em Proceedings of the IEEE/CVF Conference on Computer Vision and
  Pattern Recognition}, pages 1352--1361, 2022.

\bibitem{methane_plumes}
Elyes Ouerghi, Thibaud Ehret, Carlo de Franchis, Gabriele Facciolo, Thomas
  Lauvaux, Enric Meinhardt, and Jean-Michel Morel.
\newblock Detection of methane plumes in hyperspectral images from sentinel-5p
  by coupling anomaly detection and pattern recognition.
\newblock {\em ISPRS Annals of the Photogrammetry, Remote Sensing and Spatial
  Information Sciences}, V-3-2021:81--87, 2021.

\bibitem{methane_plumes_2}
Elyes Ouerghi, Thibaud Ehret, Carlo de Franchis, Gabriele Facciolo, Thomas
  Lauvaux, Enric Meinhardt, and Jean-Michel Morel.
\newblock Automatic methane plumes detection in time series of sentinel-5p l1b
  images.
\newblock {\em ISPRS Annals of the Photogrammetry, Remote Sensing and Spatial
  Information Sciences}, V-3-2022:147--154, 2022.

\bibitem{G-LBM}
Behnaz Rezaei, Amirreza Farnoosh, and Sarah Ostadabbas.
\newblock G-lbm: Generative low-dimensional background model estimation from
  video sequences.
\newblock In {\em European Conference on Computer Vision}, pages 293--310.
  Springer, 2020.

\bibitem{a_contrario_super_pixel}
Amandine Robin, Lionel Moisan, and Sylvie Le~Hegarat-Mascle.
\newblock An a-contrario approach for subpixel change detection in satellite
  imagery.
\newblock {\em IEEE Transactions on Pattern Analysis and Machine Intelligence},
  32(11):1977--1993, 2010.

\bibitem{siou}
Matthias Rottmann, Pascal Colling, Thomas~Paul Hack, Robin Chan, Fabian
  H{\"u}ger, Peter Schlicht, and Hanno Gottschalk.
\newblock Prediction error meta classification in semantic segmentation:
  Detection via aggregated dispersion measures of softmax probabilities.
\newblock In {\em 2020 International Joint Conference on Neural Networks
  (IJCNN)}, pages 1--9. IEEE, 2020.

\bibitem{sakkos}
Dimitrios Sakkos, Heng Liu, Jungong Han, and Ling Shao.
\newblock End-to-end video background subtraction with 3d convolutional neural
  networks.
\newblock {\em Multimedia Tools and Applications}, 77, 2018.

\bibitem{cvpr10_modelling}
Imran Saleemi, Lance Hartung, and Mubarak Shah.
\newblock Scene understanding by statistical modeling of motion patterns.
\newblock In {\em 2010 IEEE Computer Society Conference on Computer Vision and
  Pattern Recognition}, pages 2069--2076, 2010.

\bibitem{vgg}
Karen Simonyan and Andrew Zisserman.
\newblock Very deep convolutional networks for large-scale image recognition.
\newblock {\em arXiv 1409.1556}, 2014.

\bibitem{SOBRAL20144}
Andrews Sobral and Antoine Vacavant.
\newblock A comprehensive review of background subtraction algorithms evaluated
  with synthetic and real videos.
\newblock {\em Computer Vision and Image Understanding}, 122:4--21, 2014.

\bibitem{subsense}
Pierre-Luc St-Charles, Guillaume-Alexandre Bilodeau, and Robert Bergevin.
\newblock Subsense: A universal change detection method with local adaptive
  sensitivity.
\newblock {\em IEEE Transactions on Image Processing}, 24(1):359--373, 2015.

\bibitem{Stauffer_et_Grimson_vol2}
Chris Stauffer and W.E.L. Grimson.
\newblock Adaptive background mixture models for real-time tracking.
\newblock In {\em Proceedings. 1999 IEEE Computer Society Conference on
  Computer Vision and Pattern Recognition (Cat. No PR00149)}, volume~2, pages
  246--252 Vol. 2, 1999.

\bibitem{bsuv}
Mustafa~Ozan Tezcan, Prakash Ishwar, and Janusz Konrad.
\newblock Bsuv-net: A fully-convolutional neural network for background
  subtraction of unseen videos.
\newblock In {\em 2020 IEEE Winter Conference on Applications of Computer
  Vision (WACV)}, pages 2763--2772, 2020.

\bibitem{bsuv2.0}
Mustafa~Ozan Tezcan, Prakash Ishwar, and Janusz Konrad.
\newblock Bsuv-net 2.0: Spatio-temporal data augmentations for video-agnostic
  supervised background subtraction.
\newblock {\em IEEE Access}, PP:1--1, 04 2021.

\bibitem{2014cdnet}
Yi Wang, Pierre-Marc Jodoin, Fatih Porikli, Janusz Konrad, Yannick Benezeth,
  and Prakash Ishwar.
\newblock Cdnet 2014: An expanded change detection benchmark dataset.
\newblock In {\em Proceedings of the IEEE conference on computer vision and
  pattern recognition workshops}, pages 387--394, 2014.

\bibitem{witkin_1}
Andrew~P. Witkin and Jay~M. Tenenbaum.
\newblock On the role of structure in vision.
\newblock In Jacob Beck, Barbara Hope, and Azriel Rosenfeld, editors, {\em
  Human and Machine Vision}, Notes and Reports in Computer Science and Applied
  Mathematics, pages 481--543. Academic Press, 1983.

\bibitem{iccv03_paper}
Jing Zhong and Stan Sclaroff.
\newblock Segmenting foreground objects from a dynamic textured background via
  a robust kalman filter.
\newblock In {\em Proceedings Ninth IEEE International Conference on Computer
  Vision}, pages 44--50 vol.1, 2003.

\bibitem{zivkovic}
Zoran Zivkovic.
\newblock Improved adaptive gaussian mixture model for background subtraction.
\newblock In {\em Proceedings of the 17th International Conference on Pattern
  Recognition, 2004. ICPR 2004.}, volume~2, pages 28--31 Vol.2, 2004.

\end{thebibliography}
}

\clearpage

\maketitle
\thispagestyle{empty}


%
\appendix
\onecolumn
\maketitlesuppmat{Reducing false alarms in video surveillance by deep feature statistical modeling}
This supplementary section complements the description, analysis and claims of the main text. Firstly, Section~\ref{supmat_A} explains in detail the computation of the p-value, introduced in Section~\ref{sec_method}, for a mixture of Gaussians of features of even dimension. Secondly, the complete quantitative results obtained for all datasets, with and without the a-contrario validation, are provided in Section~\ref{supmat_B}. Lastly, complementary examples of the separability of TP and FP by region size versus by $logNFA$ score are provided in Section~\ref{supmat_C}.
\section{Computation of p-value for features of even dimension}
\label{supmat_A}
We seek to calculate a \textit{p}-value for a mixture of Gaussians. For simplicity, we will consider a mixture of classical Gaussian distributions without location information  
\begin{equation}\label{eq:gmm2}
    \mathbb{P}\big(\textbf{p}~|~\theta\big) = \sum_{i=1}^K \phi_i \mathbb{P}(\textbf{p}~|~\mu_i, \Sigma_i).
\end{equation}
We then define the \textit{p}-value as the integral of the density where the probability is lower than the probability density $\mathbb{P}(\textbf{p}|\theta)$:
\begin{equation}\label{eq:pvalue1}
\begin{array}{r c l}
     
    p\mbox{-}\mathrm{value}\big(\textbf{p}~|~\theta\big) & =&
    \hspace{-12mm}
    \int\limits_{\hspace{12mm}\mathcal{D}(\textbf{p}|\theta)}
    \hspace{-7mm}
    \mathbb{P}\big(\textbf{q}~|~\theta\big) \, d\textbf{q}\\
    & = & \sum_{i=1}^K \phi_i
    \hspace{-6mm}
    \int\limits_{\hspace{6mm}
    \mathcal{D}(\textbf{p}|\theta)}
    \hspace{-6mm}
    \mathbb{P}(\textbf{q}~|~\mu_i, \Sigma_i) \, d\textbf{q},
\end{array}
\end{equation}
where $\mathcal{D}(\textbf{p}|\theta) =\big\{\textbf{q}~|~\mathbb{P}(\textbf{q}|\theta )\leq \mathbb{P}(\textbf{p}|\theta)\big\}$. This yields a weighted sum of Gaussian integrals. Nevertheless, the domain $\mathcal{D}$ is impossible to compute due to the interactions between the components. 
In this situation we introduce  $\mathcal{D}_i(\textbf{p}|\theta) =\big\{\textbf{q}~|~ \phi_i \mathbb{P}(\textbf{q}~|~\mu_i, \Sigma_i) \leq \mathbb{P}(\textbf{p}|\theta)  \big\}$, which contains it ($\mathcal{D}\subseteq\mathcal{D}_i$) because if $\textbf{q}\in \mathcal{D}$ then $\forall i$ we have $ \phi_i \mathbb{P}(\textbf{q}~|~\mu_i, \Sigma_i) \leq \mathbb{P}(\textbf{q}|\theta ) \leq  \mathbb{P}(\textbf{p}|\theta) $, hence $q\in \mathcal{D}_i$. We thus find an upper bound of the \textit{p}-value by replacing $\mathcal{D}$ by the corresponding $\mathcal{D}_i$ in the integrals, i.e.
\begin{equation*}\label{eq:pvalue-upper-bound2}
    p\mbox{-}\mathrm{value}\big(\textbf{p}~|~\theta\big) \leq  
    \sum_{i=1}^K \phi_i
    \hspace{-6mm}
    \int\limits_{\hspace{6mm}
    \mathcal{D}_i(\textbf{p}|\theta)}
    \hspace{-6mm}
    \mathbb{P}(\textbf{q}~|~\mu_i, \Sigma_i) \, d\textbf{q}.
\end{equation*}
To solve this, we first rewrite the condition for a feature to be included in $D_i$, showing that this is the outside area of an ellipsoid characterized by  $R_i^2 $ as defined below:
\begin{equation}
    \begin{aligned}{r c l m}
         \phi_i \mathbb{P}(\textbf{q}~|~\mu_i, \Sigma_i) &  \leq & \mathbb{P}(\textbf{p}|\theta)  & \Leftrightarrow  \\
         \frac{\phi_i}{\sqrt{(2\pi)^d|\Sigma_i|}} \exp -\frac{1}{2}(q-\mu_i)^T\Sigma_i^{-1}(q-\mu_i) & \leq & \mathbb{P}(\textbf{p}|\theta)  & \Leftrightarrow \\
         (q-\mu_i)^T\Sigma_i^{-1}(q-\mu_i) & \geq & R_i^2 & \quad \text{where} \\
        \max (0, -2\log\mathbb{P}(\textbf{p}|\theta) + 2\log \phi_i   -\log |\Sigma_i| -d\log 2\pi )& = & R_i^2  .
    \end{aligned}
\end{equation}
We then introduce two changes of variables to facilitate the integral. The first one is a normalization of the Gaussian, $u=\Sigma_i^{-1/2}(p-\mu_i)$ with the determinant of the Jacobian $|J|=\sqrt{|\Sigma_i|}$, so that 
\begin{equation}
    \begin{array}{r c l}
        \hspace{-6mm}
    \int\limits_{\hspace{6mm}
    \mathcal{D}_i(\textbf{p}|\theta)}
    \hspace{-6mm}
    \mathbb{P}(\textbf{q}~|~\mu_i, \Sigma_i) \, d\textbf{q}& = &  
    (2\pi)^{-d/2}
    \hspace{-6mm}
    \int\limits_{\hspace{6mm}
    u^Tu\geq R_i^2}
    \hspace{-6mm}
    e^{-\frac{1}{2}u^Tu}\, du.
    \end{array}
\end{equation}
Subsequently, we shift to hyper spherical coordinates through another change of variable:
\begin{equation}
    \begin{array}{r l}
        u_1 & = r \cos{\theta_1}  \\
        u_2 & = r \sin{\theta_1}\cos{\theta_2}\\
        \vdots & = \vdots \\
        u_{d-2} & = r \sin{\theta_1}\hdots \sin{\theta_{d-2}}\cos{\theta_{d-1}}\\
        u_{d-1} & = r \sin{\theta_1}\hdots \sin{\theta_{d-2}}\sin{\theta_{d-1}}.
    \end{array}
\end{equation}
The determinant of the Jacobian of this change is $|J|=r^{d-1}\prod_{k=1}^{d-2}\sin^{d-k-1}\theta_k$. Thus, we have the integral
\begin{equation}
    \begin{array}{r c l}
    (2\pi)^{-d/2}
    \hspace{-6mm}
    \int\limits_{\hspace{6mm}
    u^Tu\geq R_i^2}
    \hspace{-6mm}
     e^{-\frac{1}{2}u^Tu}\, du& =& (2\pi)^{-d/2} \int_0^{2\pi}d\theta_d \prod_{k=1}^{d-2}\int_0^\pi\sin^{d-k-1}\theta_kd\theta_k\int_{R_i}^{+\infty}r^{d-1}\exp -\frac{1}{2}r^2\, dr.
    \end{array}
\end{equation}
We first integrate the angles 
\begin{equation}
    \begin{array}{r c l}
        \prod_{k=1}^{d-2}\int_0^\pi\sin^{d-k-1}\theta_kd\theta_k & = & 2\frac{\pi^{d/2}}{\Gamma(d/2)}.
    \end{array}
\end{equation}
Then, we recognize the probability density of $\chi^2$ in the integral and therefore the whole is the survival function of $\chi^2$, such that
\begin{equation}
    \begin{array}{r c l}
        \hspace{-6mm}
    \int\limits_{\hspace{6mm}
    \mathcal{D}_i(\textbf{p}|\theta)}
    \hspace{-6mm}
    \mathbb{P}(\textbf{q}~|~\mu_i, \Sigma_i) \, d\textbf{q}& = &  
    \int_{R_i}^{+\infty} \frac{1}{2^{d/2 -1}\Gamma(d/2)}r^{d-1}e^{-\frac{1}{2}r^2}\, dr \\
    & =&\int_{R_i^2}^{+\infty} \frac{1}{2^{d/2}\Gamma(d/2)}x^{d/2 - 1}e^{-\frac{1}{2}x}\, dx \\ 
    & = & \textbf{SF}_{\chi^2} (R_i^2).
    \end{array}
\end{equation}
The $r$ integral can be computed via integration by parts, and the result depends on whether the dimension $d$ of the features is even or odd. We will first see the even case $d=2m$:
\begin{equation}
    \begin{array}{r c l}
     
    \int_{R_i}^{+\infty} r^{2m-1}e^{-\frac{1}{2}r^2}\, dr & = & R_i^{2m-2}e^{-\frac{1}{2}R_i^2}+(2m-2)\int_{R_i}^{+\infty} r^{2m-3}e^{-\frac{1}{2}r^2}\, dr \\
    &=& R_i^{2m-2}e^{-\frac{1}{2}R_i^2}+(2m-2)\left[ R_i^{2m-4}e^{-\frac{1}{2}R_i^2}+(2m-4)\int_{R_i}^{+\infty} r^{2m-5}e^{-\frac{1}{2}r^2}\, dr\right] \\
    &\vdots & \\
    & = & \sum_{j=1}^{m-1}\frac{(m-1)!}{j!}2^{m-1-j}R_i^{2j}e^{-\frac{1}{2}R_i^2} + 2^{m-1}(m-1)!\int_{R_i}^{+\infty} re^{-\frac{1}{2}r^2} \\
    & = & \sum_{j=1}^{m-1}\frac{(m-1)!}{j!}2^{m-1-j}R_i^{2j}e^{-\frac{1}{2}R_i^2} + 2^{m-1}(m-1)!e^{-\frac{1}{2}R_i^2} \\
    & = & \sum_{j=0}^{m-1}\frac{(m-1)!}{j!}2^{m-1-j}R_i^{2j}e^{-\frac{1}{2}R_i^2}.
    \end{array}
\end{equation}
We proceed in the same way for the odd case $d=2m+1$:
\begin{equation}
    \begin{array}{r c l}
     
    \int_{R_i}^{+\infty} r^{2m}e^{-\frac{1}{2}r^2}\, dr & = & R_i^{2m-1}e^{-\frac{1}{2}R_i^2}+(2m-1)\int_{R_i}^{+\infty} r^{2m-2}e^{-\frac{1}{2}r^2}\, dr \\
    &=& R_i^{2m-1}e^{-\frac{1}{2}R_i^2}+(2m-1)\left[ R_i^{2m-3}e^{-\frac{1}{2}R_i^2}+(2m-3)\int_{R_i}^{+\infty} r^{2m-4}e^{-\frac{1}{2}r^2}\, dr\right] \\
    &\vdots & \\
    & = & \sum_{j=1}^{m}\frac{(2m)!j!}{m!(2j!)}R_i^{2j-1}e^{-\frac{1}{2}R_i^2} + \frac{(2m)!}{2(m!)}\int_{R_i}^{+\infty}e^{-\frac{1}{2}r^2} \,dr \\
    & = & \sum_{j=1}^{m}\frac{(2m)!j!}{m!(2j!)}R_i^{2j-1}e^{-\frac{1}{2}R_i^2} + \frac{(2m)!}{2(m!)}\sqrt{\frac{\pi}{2}}\left[1-\textbf{erf}(R_i/\sqrt{2})) \right].\\
    \end{array}
\end{equation}
The advantage of the even case is that it takes the form of a finite sum that is simple to compute, whereas the odd case requires estimating the error function. We get the complete formula of the even case by combining 17 and 16
\begin{equation}
p\mbox{-}\mathrm{value}\big(\textbf{p}~|~\theta\big) \leq \sum_{i=1}^K\phi_i \sum_{j=0}^{m-1}\frac{2^{-j}}{j!}R_i^{2j}e^{-\frac{1}{2}R_i^2}.
\end{equation}
We get the complete formula of the odd case by combining 18 and 16
\begin{equation}
p\mbox{-}\mathrm{value}\big(\textbf{p}~|~\theta\big) \leq \sum_{i=1}^K\phi_i \left(\sum_{j=1}^{m}2^m\sqrt{\frac{2}{\pi}}\frac{j!}{(2j)!}R_i^{2j-1}e^{-\frac{1}{2}R_i^2} + 2^{m-1}\left[1-\textbf{erf}(R_i/\sqrt{2})) \right]\right).
\end{equation}
\newpage
\section{Quantitative results}
\label{supmat_B}
This section provides the results with the raw metric values obtained in our experiments. In addition to the object-wise metrics defined in the main text, we provide a complementary set of object-level metrics analogous to the pixel-level ones. The sIoU and $\fe^{sIoU}$ metrics address the case of fragmented detections. Nevertheless, they are based on a measurement that might consider as faulty detections those that do not spatially align with the ground truth to a degree. While this can be considered valid for most cases, sometimes the nature of the target application might not allow us to miss any detections, even if the precision of the blob is quite dissimilar from the ground truth. Hence, we adapt the pixel-level metrics to the object level where any detection which overlaps by at least one pixel with the ground truth will be considered as a good detection. 

We thus define $\tp^{ob}$, $\fn^{ob}$ and $\fp^{ob}$ as true positives, false negatives and false positives for sets of connected components. The true negatives, however, lack a clear meaning. Alternatively, we propose to compute the \fpr relative to the number of frames $n_f$. Instead of selecting $\tp^{ob}$, $\fn^{ob}$ and $\fp^{ob}$ by IoU thresholding, we define $\tp^{ob}$ as detected regions containing at least one real positive pixel. Then,  $\fp^{ob}$ are detections that do not overlap with any real positive pixels and $\fn^{ob}$ are regions that should have been detected but no change was computed by the method. To avoid the fragmentation issue, we carefully mark the entire ground truth region for each true positive detection as already checked. Thus, we define the object-wise metrics as follows:
\begin{itemize}
\itemsep0em
  \item $\pr^{ob} = \tp^{ob} / (\tp^{ob} + \fp^{ob})$
  \item $\re^{ob} = \tp^{ob} / (\tp^{ob}+\fn^{ob})$
  \item $\fpr^{ob} = \fp^{ob}/n_f$
  \item $\pwc^{ob} = 100 \times (\fn^{ob}+\fp^{ob}) / (\tp^{ob}+\fn^{ob}+\fp^{ob})$
  \item $\fe^{ob} = 2 (\pr^{ob} \times \re^{ob}) / (\pr^{ob} + \re^{ob})$.
\end{itemize}
 These object-wise metrics are included in the raw tables provided in this section.

\subsection{CDNet dataset}
Table~\ref{table:cdnet_overall_results} shows the overall results obtained for each evaluation metric, before and after the a-contrario validation, considering all sequences and all categories of the CDNet benchmark. Notice that, since we assume the camera is static, the categories \textit{cameraJitter} and \textit{PTZ} have not been considered. Additionally, the results for each category are provided separately. Each category is linked to its corresponding table in the list down below:
\begin{itemize}
\itemsep0em
  \item \textit{baseline}: Table~\ref{table:results_cdnet_baseline}
  \item \textit{dynamicBackground}: Table~\ref{table:results_cdnet_dynamicbackground}
  \item \textit{badWeather}: Table~\ref{table:results_cdnet_badweather}
  \item \textit{intermittentObjectMotion}: Table~\ref{table:results_cdnet_intermittentobjectmotion}
  \item \textit{lowFramerate}: Table~\ref{table:results_cdnet_lowframerate}
  \item \textit{nightVideos}: Table~\ref{table:results_cdnet_nightvideos}
  \item \textit{thermal}: Table~\ref{table:results_cdnet_thermal}
  \item \textit{shadow}: Table~\ref{table:results_cdnet_shadow}
  \item \textit{turbulence}: Table~\ref{table:results_cdnet_turbulence}
\end{itemize}

\renewcommand{\arraystretch}{1.2}
\begin{table*}[t]
\centering
\resizebox{1\textwidth}{!}{
\begin{tabular}{lccccccccccccccc}
Method                      & a contrario validation & & \multicolumn{5}{c}{pixel-wise} & \multicolumn{7}{c}{object-wise} \\

                            \cmidrule[0.4pt](lr{0.125em}){3-7}%
                            \cmidrule[0.4pt](lr{0.125em}){8-16}%
                            &                        & \re$^{pi}\uparrow$    & \fpr$^{pi}\downarrow$   & \pwc$^{pi}\downarrow$   & \pr$^{pi}\uparrow$    & \fe$^{pi}\uparrow$   & \re$^{ob}\uparrow$     & \fpr$^{ob}\downarrow$     & \pwc$^{ob}\downarrow$    & \pr$^{ob}\uparrow$     & \fe$^{ob}\uparrow$   & $sIoU\uparrow$ & \fe$^{sIoU}\uparrow$ \\

                            \cmidrule[0.4pt](lr{0.125em}){3-3}%
                            \cmidrule[0.4pt](lr{0.125em}){4-4}%
                            \cmidrule[0.4pt](lr{0.125em}){5-5}%
                            \cmidrule[0.4pt](lr{0.125em}){6-6}%
                            \cmidrule[0.4pt](lr{0.125em}){7-7}%
                            \cmidrule[0.4pt](lr{0.125em}){8-8}%
                            \cmidrule[0.4pt](lr{0.125em}){9-9}%
                            \cmidrule[0.4pt](lr{0.125em}){10-10}%
                            \cmidrule[0.4pt](lr{0.125em}){11-11}%
                            \cmidrule[0.4pt](lr{0.125em}){12-12}%
                            \cmidrule[0.4pt](lr{0.125em}){13-13}%
                            \cmidrule[0.4pt](lr{0.125em}){14-14}%
                            \cmidrule[0.4pt](lr{0.125em}){15-15}%
                            \cmidrule[0.4pt](lr{0.125em}){16-16}%

ViBe~\cite{vibe}                              & \cross                                              & \textbf{0.542}  & 0.01032   & 2.7937   & 0.717  & 0.538  & \textbf{0.879}   & 49.296    & 93.452   & 0.068   & 0.116   & 0.019      & 0.182    \\
\textbf{}                                                                     & \checkmark          & 0.519  & \textbf{0.00367}   & \textbf{2.1565}   & \textbf{0.829}  & 0.\textbf{610}  & 0.801   & \textbf{2.980}     & \textbf{77.457}   & \textbf{0.255}   & \textbf{0.342}   & \textbf{0.056}      & \textbf{0.248}    \\ \hline
SuBSENSE~\cite{subsense}                      & \cross                                              & \textbf{0.775}  & 0.00593   & 1.4354   & 0.803  & 0.\textbf{760}  & \textbf{0.810}   & 0.633     & 52.826   & 0.550   & 0.607   & 0.243      & 0.348    \\
\textbf{}                                                                     & \checkmark          & 0.735  & \textbf{0.00472}   & \textbf{1.3915}   & \textbf{0.818}  & 0.751  & 0.761   & \textbf{}0.407     & \textbf{48.903}   & \textbf{0.622}   & \textbf{0.642}   & \textbf{0.280 }     & \textbf{0.378}    \\ \hline
SemanticBGS (with ViBe)~\cite{semanticbgs}      & \cross                                            & \textbf{0.624}  & 0.00360   & 1.6006   & 0.806  & 0.658  & \textbf{0.859}   & 17.703    & 89.510   & 0.109   & 0.172   & 0.053      & 0.109    \\
\textbf{}                                                                     & \checkmark          & 0.602  & \textbf{0.00212}   & \textbf{1.4934}   & \textbf{0.889}  & 0\textbf{.684}  & 0.783   & \textbf{2.024}     & \textbf{70.724}   & \textbf{0.323}   & \textbf{0.408}   & \textbf{0.116}      & \textbf{0.212}    \\ \hline
SemanticBGS (with SuBSENSE)~\cite{semanticbgs}  & \cross                                            & \textbf{0.776}  & 0.00406   & 1.1590   & 0.848  & \textbf{0.787}  & \textbf{0.794}   & 1.778     & 54.892   & 0.528   & 0.578   & 0.228      & 0.330    \\
\textbf{}                                                                     & \checkmark          & 0.738  & \textbf{0.00329}   & \textbf{1.1573}   & \textbf{0.861}  & 0.773  & 0.746   & \textbf{0.360}     & \textbf{47.472}   & \textbf{0.649}   & \textbf{0.652}   & \textbf{0.298}      & \textbf{0.395}    \\ \hline
BSUV-Net~\cite{bsuv}                           & \cross                                             & \textbf{0.784}  & 0.00238   & \textbf{0.6373}   & 0.880  & \textbf{0.801}  & \textbf{0.745}   & 0.471     & 49.858   & 0.600   & 0.625   & 0.317      & 0.438    \\
\textbf{}                                                                     & \checkmark          & 0.761  & \textbf{0.00210}   &  0.6620  & \textbf{0.909}  & 0.788  & 0.712   & \textbf{0.279}     & \textbf{45.999}   & \textbf{0.694}   & \textbf{0.659}   &\textbf{ 0.369}      & \textbf{0.469}    \\ \hline
BSUV-Net 2.0~\cite{bsuv2.0}                   & \cross                                              & \textbf{0.683}  & \textbf{0.00124}   & \textbf{0.7886}   & 0.945  & \textbf{0.755}  & \textbf{0.641 }  & 0.279     & 52.992   & 0.637   & 0.603   & 0.312      & 0.417    \\
\textbf{}                                                                     & \checkmark          & 0.676  & 0.00127   &  0.8045  & \textbf{0.949}  & 0.749  & 0.625   & \textbf{0.150 }    & \textbf{47.985}   & \textbf{0.772}   & \textbf{0.644}   & \textbf{0.390}      & \textbf{0.456}    \\ \hline
BSUV-Net FPM~\cite{bsuv}                    & \cross                                                & \textbf{0.665}  & 0.00155   & \textbf{0.8659}   & 0.911  & \textbf{0.724}  & \textbf{0.671}   & 0.246     & 49.390   & 0.691   & 0.634   & 0.357      & 0.444    \\
\textbf{}                                                                     & \checkmark          & 0.654  & 0.\textbf{00150}   &  0.8764  & \textbf{0.913}  & 0.715  & 0.650   & \textbf{0.162}     & \textbf{46.671}   & \textbf{0.771}   & \textbf{0.655}   & \textbf{0.391}      & \textbf{0.458}    \\ \hline
\end{tabular}
}
\caption{Average pixel-wise and object-wise metrics for all evaluated categories of the CDNet dataset. The categories \textit{cameraJitter} and \textit{PTZ} have been left out as they do not involve a static camera.}
\label{table:cdnet_overall_results}
\end{table*}

\begin{table*}[t]
\centering
\resizebox{1\textwidth}{!}{
\begin{tabular}{lccccccccccccccc}
\hline
Method                      & a contrario validation & & \multicolumn{5}{c}{pixel-wise} & \multicolumn{7}{c}{object-wise} \\

                            \cmidrule[0.4pt](lr{0.125em}){3-7}%
                            \cmidrule[0.4pt](lr{0.125em}){8-16}%
                            &                        & \re$^{pi}\uparrow$    & \fpr$^{pi}\downarrow$   & \pwc$^{pi}\downarrow$   & \pr$^{pi}\uparrow$    & \fe$^{pi}\uparrow$   & \re$^{ob}\uparrow$     & \fpr$^{ob}\downarrow$     & \pwc$^{ob}\downarrow$    & \pr$^{ob}\uparrow$     & \fe$^{ob}\uparrow$   & $sIoU\uparrow$ & \fe$^{sIoU}\uparrow$ \\

                            \cmidrule[0.4pt](lr{0.125em}){3-3}%
                            \cmidrule[0.4pt](lr{0.125em}){4-4}%
                            \cmidrule[0.4pt](lr{0.125em}){5-5}%
                            \cmidrule[0.4pt](lr{0.125em}){6-6}%
                            \cmidrule[0.4pt](lr{0.125em}){7-7}%
                            \cmidrule[0.4pt](lr{0.125em}){8-8}%
                            \cmidrule[0.4pt](lr{0.125em}){9-9}%
                            \cmidrule[0.4pt](lr{0.125em}){10-10}%
                            \cmidrule[0.4pt](lr{0.125em}){11-11}%
                            \cmidrule[0.4pt](lr{0.125em}){12-12}%
                            \cmidrule[0.4pt](lr{0.125em}){13-13}%
                            \cmidrule[0.4pt](lr{0.125em}){14-14}%
                            \cmidrule[0.4pt](lr{0.125em}){15-15}%
                            \cmidrule[0.4pt](lr{0.125em}){16-16}%
ViBe~\cite{vibe}                              & \cross       & 0.675 & 0.0005 & 1.413 & 0.968 & 0.790 & 0.913 & 10.915 & 83.091 & 0.173 & 0.280 & 0.069 & 0.148 \\
\textbf{}                                 & \checkmark       & 0.669 & 0.0004 & 1.416 & 0.970 & 0.787 & 0.906 & 2.612 & 56.979 & 0.465 & 0.580 & 0.117 & 0.256 \\ \hline
SuBSENSE~\cite{subsense}                      & \cross       & 0.946 & 0.0017 & 0.344 & 0.940 & 0.942 & 0.846 & 0.101 & 21.209 & 0.915 & 0.877 & 0.530 & 0.684 \\
                                              & \checkmark   & 0.946 & 0.0017 & 0.344 & 0.941 & 0.942 & 0.844 & 0.083 & 20.650 & 0.927 & 0.881 & 0.537 & 0.687 \\ \hline
SemanticBGS (with ViBe)~\cite{semanticbgs}      & \cross     & 0.908 & 0.0012 & 0.395 & 0.967 & 0.936 & 0.911 & 4.130 & 68.666 & 0.329 & 0.463 & 0.162 & 0.344 \\
                                                & \checkmark & 0.906 & 0.0011 & 0.392 & 0.968 & 0.936 & 0.904 & 1.239 & 43.840 & 0.610 & 0.711 & 0.270 & 0.504 \\ \hline
SemanticBGS (with SuBSENSE)~\cite{semanticbgs}  & \cross     & 0.963 & 0.0024 & 0.328 & 0.935 & 0.948 & 0.850 & 0.150 & 22.802 & 0.896 & 0.870 & 0.542 & 0.684 \\
                                                & \checkmark & 0.963 & 0.0023 & 0.325 & 0.935 & 0.948 & 0.846 & 0.019 & 16.845 & 0.980 & 0.907 & 0.594 & 0.716 \\ \hline
BSUV-Net~\cite{bsuv}                           & \cross      & 0.963 & 0.0008 & 0.176 & 0.962 & 0.963 & 0.868 & 0.079 & 16.930 & 0.951 & 0.907 & 0.647 & 0.782 \\
& \checkmark                                                 & 0.963 & 0.0008 & 0.176 & 0.962 & 0.962 & 0.864 & 0.033 & 15.015 & 0.983 & 0.919 & 0.673 & 0.792 \\ \hline
BSUV-Net 2.0~\cite{bsuv2.0}                   & \cross       & 0.924 & 0.0004 & 0.247 & 0.980 & 0.951 & 0.777 & 0.077 & 26.040 & 0.943 & 0.848 & 0.623 & 0.765 \\
& \checkmark                                                 & 0.923 & 0.0004 & 0.247 & 0.980 & 0.950 & 0.768 & 0.017 & 24.187 & 0.985 & 0.860 & 0.656 & 0.770 \\ \hline
BSUV-Net FPM~\cite{bsuv}                    & \cross         & 0.934 & 0.0006 & 0.198 & 0.976 & 0.954 & 0.812 & 0.020 & 19.990 & 0.983 & 0.886 & 0.658 & 0.796 \\
& \checkmark                                                 & 0.934 & 0.0006 & 0.198 & 0.976 & 0.954 & 0.809 & 0.005 & 19.391 & 0.996 & 0.890 & 0.670 & 0.797 \\ \hline
\end{tabular}
}
\caption{Average pixel-wise and object-wise metrics for the \textit{baseline} category of the CDNet dataset.}
\label{table:results_cdnet_baseline}
\end{table*}

\begin{table*}[t]
\centering
\resizebox{1\textwidth}{!}{
\begin{tabular}{lccccccccccccccc}
\hline
Method                      & a contrario validation & & \multicolumn{5}{c}{pixel-wise} & \multicolumn{7}{c}{object-wise} \\

                            \cmidrule[0.4pt](lr{0.125em}){3-7}%
                            \cmidrule[0.4pt](lr{0.125em}){8-16}%
                            &                        & \re$^{pi}\uparrow$    & \fpr$^{pi}\downarrow$   & \pwc$^{pi}\downarrow$   & \pr$^{pi}\uparrow$    & \fe$^{pi}\uparrow$   & \re$^{ob}\uparrow$     & \fpr$^{ob}\downarrow$     & \pwc$^{ob}\downarrow$    & \pr$^{ob}\uparrow$     & \fe$^{ob}\uparrow$   & $sIoU\uparrow$ & \fe$^{sIoU}\uparrow$ \\

                            \cmidrule[0.4pt](lr{0.125em}){3-3}%
                            \cmidrule[0.4pt](lr{0.125em}){4-4}%
                            \cmidrule[0.4pt](lr{0.125em}){5-5}%
                            \cmidrule[0.4pt](lr{0.125em}){6-6}%
                            \cmidrule[0.4pt](lr{0.125em}){7-7}%
                            \cmidrule[0.4pt](lr{0.125em}){8-8}%
                            \cmidrule[0.4pt](lr{0.125em}){9-9}%
                            \cmidrule[0.4pt](lr{0.125em}){10-10}%
                            \cmidrule[0.4pt](lr{0.125em}){11-11}%
                            \cmidrule[0.4pt](lr{0.125em}){12-12}%
                            \cmidrule[0.4pt](lr{0.125em}){13-13}%
                            \cmidrule[0.4pt](lr{0.125em}){14-14}%
                            \cmidrule[0.4pt](lr{0.125em}){15-15}%
                            \cmidrule[0.4pt](lr{0.125em}){16-16}%
ViBe~\cite{vibe}                              & \cross    &   0.876 & 0.054 & 5.354 & 0.173 & 0.273 & 0.895 & 302.477 & 99.833 & 0.002 & 0.003 & 0.000 & 0.001 \\
  & \checkmark   &   0.863 & 0.006 & 0.371 & 0.624 & 0.707 & 0.850 & 4.696 & 92.703 & 0.075 & 0.132 & 0.018 & 0.038 \\ \hline
SuBSENSE~\cite{subsense}                      & \cross       & 0.825 & 0.000 & 0.265 & 0.914 & 0.859 & 0.766 & 0.183 & 49.715 & 0.614 & 0.654 & 0.224 & 0.348 \\
  & \checkmark   &  0.816 & 0.000 & 0.257 & 0.942 & 0.869 & 0.742 & 0.051 & 35.901 & 0.826 & 0.774 & 0.309 & 0.447 \\ \hline
SemanticBGS (with ViBe)~\cite{semanticbgs}      & \cross     & 0.919 & 0.007 & 0.739 & 0.578 & 0.668 & 0.889 & 50.017 & 98.828 & 0.012 & 0.023 & 0.003 & 0.006 \\
  & \checkmark   &  0.910 & 0.001 & 0.168 & 0.850 & 0.878 & 0.847 & 1.550 & 72.911 & 0.177 & 0.279 & 0.048 & 0.096 \\ \hline
SemanticBGS (with SuBSENSE)~\cite{semanticbgs}  & \cross     & 0.935 & 0.001 & 0.114 & 0.947 & 0.941 & 0.773 & 0.061 & 34.886 & 0.816 & 0.781 & 0.325 & 0.457 \\
  & \checkmark   &  0.931 & 0.001 & 0.114 & 0.952 & 0.941 & 0.751 & 0.021 & 30.075 & 0.906 & 0.817 & 0.422 & 0.557 \\ \hline
BSUV-Net~\cite{bsuv}                           & \cross      & 0.639 & 0.000 & 0.236 & 0.959 & 0.720 & 0.546 & 0.114 & 55.167 & 0.718 & 0.601 & 0.206 & 0.337 \\
  & \checkmark  &   0.639 & 0.000 & 0.234 & 0.967 & 0.719 & 0.534 & 0.009 & 47.751 & 0.958 & 0.669 & 0.269 & 0.393 \\ \hline
BSUV-Net 2.0~\cite{bsuv2.0}                   & \cross       & 0.738 & 0.002 & 0.256 & 0.897 & 0.771 & 0.599 & 0.064 & 48.351 & 0.786 & 0.661 & 0.260 & 0.366 \\
  & \checkmark  &   0.740 & 0.002 & 0.302 & 0.921 & 0.766 & 0.610 & 0.006 & 39.987 & 0.976 & 0.726 & 0.389 & 0.487 \\ \hline
BSUV-Net FPM~\cite{bsuv}                    & \cross         & 0.549 & 0.001 & 0.561 & 0.925 & 0.646 & 0.562 & 0.088 & 53.823 & 0.737 & 0.611 & 0.415 & 0.500 \\
  & \checkmark  &   0.546 & 0.001 & 0.562 & 0.927 & 0.644 & 0.550 & 0.015 & 47.132 & 0.935 & 0.668 & 0.450 & 0.514 \\ \hline
\end{tabular}
}
\caption{Average pixel-wise and object-wise metrics for the \textit{dynamicBackground} category of the CDNet dataset.}
\label{table:results_cdnet_dynamicbackground}
\end{table*}

\begin{table*}[t]
\centering
\resizebox{1\textwidth}{!}{
\begin{tabular}{lccccccccccccccc}
\hline
Method                      & a contrario validation & & \multicolumn{5}{c}{pixel-wise} & \multicolumn{7}{c}{object-wise} \\

                            \cmidrule[0.4pt](lr{0.125em}){3-7}%
                            \cmidrule[0.4pt](lr{0.125em}){8-16}%
                            &                        & \re$^{pi}\uparrow$    & \fpr$^{pi}\downarrow$   & \pwc$^{pi}\downarrow$   & \pr$^{pi}\uparrow$    & \fe$^{pi}\uparrow$   & \re$^{ob}\uparrow$     & \fpr$^{ob}\downarrow$     & \pwc$^{ob}\downarrow$    & \pr$^{ob}\uparrow$     & \fe$^{ob}\uparrow$   & $sIoU\uparrow$ & \fe$^{sIoU}\uparrow$ \\

                            \cmidrule[0.4pt](lr{0.125em}){3-3}%
                            \cmidrule[0.4pt](lr{0.125em}){4-4}%
                            \cmidrule[0.4pt](lr{0.125em}){5-5}%
                            \cmidrule[0.4pt](lr{0.125em}){6-6}%
                            \cmidrule[0.4pt](lr{0.125em}){7-7}%
                            \cmidrule[0.4pt](lr{0.125em}){8-8}%
                            \cmidrule[0.4pt](lr{0.125em}){9-9}%
                            \cmidrule[0.4pt](lr{0.125em}){10-10}%
                            \cmidrule[0.4pt](lr{0.125em}){11-11}%
                            \cmidrule[0.4pt](lr{0.125em}){12-12}%
                            \cmidrule[0.4pt](lr{0.125em}){13-13}%
                            \cmidrule[0.4pt](lr{0.125em}){14-14}%
                            \cmidrule[0.4pt](lr{0.125em}){15-15}%
                            \cmidrule[0.4pt](lr{0.125em}){16-16}%
ViBe~\cite{vibe}                              & \cross    & 0.494 & 0.002 & 1.000 & 0.897 & 0.622 & 0.879 & 10.514 & 94.330 & 0.057 & 0.104 & 0.012 & 0.024 \\
 & \checkmark   & 0.483 & 0.000 & 0.866 & 0.917 & 0.759 & 0.801 & 1.501 & 80.507 & 0.212 & 0.319 & 0.047 & 0.102 \\ \hline
SuBSENSE~\cite{subsense}                      & \cross       & 0.783 & 0.001 & 0.507 & 0.929 & 0.846 & 0.772 & 0.426 & 59.518 & 0.468 & 0.559 & 0.260 & 0.392 \\
& \checkmark   & 0.781 & 0.001 & 0.501 & 0.941 & 0.849 & 0.765 & 0.300 & 53.159 & 0.550 & 0.628 & 0.318 & 0.453 \\ \hline
SemanticBGS (with ViBe)~\cite{semanticbgs}      & \cross     & 0.561 & 0.000 & 0.809 & 0.962 & 0.693 & 0.839 & 6.863 & 93.027 & 0.071 & 0.127 & 0.019 & 0.041 \\
& \checkmark   & 0.554 & 0.000 & 0.804 & 0.983 & 0.690 & 0.765 & 1.240 & 78.923 & 0.233 & 0.344 & 0.063 & 0.135 \\ \hline
SemanticBGS (with SuBSENSE)~\cite{semanticbgs}  & \cross     & 0.764 & 0.001 & 0.561 & 0.947 & 0.839 & 0.762 & 2.596 & 71.185 & 0.343 & 0.414 & 0.184 & 0.274 \\
& \checkmark   & 0.762 & 0.001 & 0.558 & 0.952 & 0.840 & 0.754 & 0.300 & 54.231 & 0.540 & 0.619 & 0.298 & 0.431 \\ \hline
BSUV-Net~\cite{bsuv}                           & \cross      & 0.659 & 0.000 & 0.574 & 0.973 & 0.771 & 0.449 & 0.317 & 75.486 & 0.380 & 0.386 & 0.282 & 0.430 \\
& \checkmark   & 0.658 & 0.000 & 0.577 & 0.974 & 0.770 & 0.436 & 0.188 & 71.281 & 0.497 & 0.434 & 0.361 & 0.485 \\ \hline
BSUV-Net 2.0~\cite{bsuv2.0}                   & \cross       & 0.667 & 0.001 & 0.631 & 0.956 & 0.757 & 0.545 & 0.228 & 64.654 & 0.495 & 0.504 & 0.236 & 0.319 \\
& \checkmark   & 0.666 & 0.001 & 0.632 & 0.956 & 0.756 & 0.536 & 0.173 & 62.841 & 0.540 & 0.521 & 0.315 & 0.360 \\ \hline
BSUV-Net FPM~\cite{bsuv}                    & \cross         & 0.793 & 0.001 & 0.417 & 0.965 & 0.867 & 0.685 & 0.334 & 61.731 & 0.477 & 0.552 & 0.287 & 0.387 \\
& \checkmark   & 0.790 & 0.001 & 0.418 & 0.966 & 0.866 & 0.672 & 0.200 & 55.300 & 0.590 & 0.615 & 0.322 & 0.409 \\ \hline
\end{tabular}
}
\caption{Average pixel-wise and object-wise metrics for the \textit{badWeather} category of the CDNet dataset.}
\label{table:results_cdnet_badweather}
\end{table*}

\begin{table*}[t]
\centering
\resizebox{1\textwidth}{!}{
\begin{tabular}{lccccccccccccccc}
\hline
Method                      & a contrario validation & & \multicolumn{5}{c}{pixel-wise} & \multicolumn{7}{c}{object-wise} \\

                            \cmidrule[0.4pt](lr{0.125em}){3-7}%
                            \cmidrule[0.4pt](lr{0.125em}){8-16}%
                            &                        & \re$^{pi}\uparrow$    & \fpr$^{pi}\downarrow$   & \pwc$^{pi}\downarrow$   & \pr$^{pi}\uparrow$    & \fe$^{pi}\uparrow$   & \re$^{ob}\uparrow$     & \fpr$^{ob}\downarrow$     & \pwc$^{ob}\downarrow$    & \pr$^{ob}\uparrow$     & \fe$^{ob}\uparrow$   & $sIoU\uparrow$ & \fe$^{sIoU}\uparrow$ \\

                            \cmidrule[0.4pt](lr{0.125em}){3-3}%
                            \cmidrule[0.4pt](lr{0.125em}){4-4}%
                            \cmidrule[0.4pt](lr{0.125em}){5-5}%
                            \cmidrule[0.4pt](lr{0.125em}){6-6}%
                            \cmidrule[0.4pt](lr{0.125em}){7-7}%
                            \cmidrule[0.4pt](lr{0.125em}){8-8}%
                            \cmidrule[0.4pt](lr{0.125em}){9-9}%
                            \cmidrule[0.4pt](lr{0.125em}){10-10}%
                            \cmidrule[0.4pt](lr{0.125em}){11-11}%
                            \cmidrule[0.4pt](lr{0.125em}){12-12}%
                            \cmidrule[0.4pt](lr{0.125em}){13-13}%
                            \cmidrule[0.4pt](lr{0.125em}){14-14}%
                            \cmidrule[0.4pt](lr{0.125em}){15-15}%
                            \cmidrule[0.4pt](lr{0.125em}){16-16}%
ViBe~\cite{vibe}                              & \cross    & 0.354 & 0.020 & 6.686 & 0.693 & 0.412 & 0.761 & 6.246 & 88.585 & 0.123 & 0.200 & 0.012 & 1.272 \\
& \checkmark   & 0.343 & 0.019 & 6.716 & 0.711 & 0.398 & 0.743 & 2.315 & 74.249 & 0.294 & 0.392 & 0.023 & 1.243 \\ \hline
SuBSENSE~\cite{subsense}                      & \cross       & 0.592 & 0.008 & 4.476 & 0.773 & 0.613 & 0.735 & 0.297 & 42.834 & 0.726 & 0.696 & 0.135 & 0.180 \\
& \checkmark   & 0.588 & 0.008 & 4.499 & 0.775 & 0.610 & 0.720 & 0.248 & 42.453 & 0.750 & 0.700 & 0.144 & 0.189 \\ \hline
SemanticBGS (with ViBe)~\cite{semanticbgs}      & \cross     & 0.486 & 0.009 & 4.855 & 0.838 & 0.612 & 0.781 & 3.639 & 84.991 & 0.161 & 0.257 & 0.141 & 0.212 \\
& \checkmark   & 0.476 & 0.009 & 4.900 & 0.846 & 0.605 & 0.763 & 1.308 & 67.380 & 0.368 & 0.472 & 0.178 & 0.242 \\ \hline
SemanticBGS (with SuBSENSE)~\cite{semanticbgs}  & \cross     & 0.651 & 0.005 & 3.670 & 0.857 & 0.721 & 0.739 & 0.433 & 47.370 & 0.648 & 0.661 & 0.024 & 0.057 \\
& \checkmark   & 0.647 & 0.005 & 3.703 & 0.858 & 0.718 & 0.723 & 0.237 & 42.061 & 0.747 & 0.704 & 0.048 & 0.096 \\ \hline
BSUV-Net~\cite{bsuv}                           & \cross      & 0.866 & 0.003 & 1.186 & 0.968 & 0.911 & 0.886 & 0.055 & 17.878 & 0.916 & 0.900 & 0.317 & 0.430 \\
& \checkmark   & 0.866 & 0.003 & 1.187 & 0.968 & 0.911 & 0.884 & 0.037 & 15.968 & 0.942 & 0.912 & 0.338 & 0.448 \\ \hline
BSUV-Net 2.0~\cite{bsuv2.0}                   & \cross       & 0.725 & 0.003 & 1.786 & 0.975 & 0.823 & 0.831 & 0.182 & 29.605 & 0.825 & 0.825 & 0.217 & 0.346 \\
& \checkmark   & 0.723 & 0.003 & 1.801 & 0.975 & 0.821 & 0.822 & 0.057 & 22.762 & 0.932 & 0.871 & 0.264 & 0.381 \\ \hline
BSUV-Net FPM~\cite{bsuv}                    & \cross         & 0.762 & 0.003 & 1.562 & 0.978 & 0.854 & 0.861 & 0.050 & 18.578 & 0.938 & 0.896 & 0.331 & 0.461 \\
& \checkmark   & 0.762 & 0.003 & 1.565 & 0.978 & 0.854 & 0.857 & 0.031 & 17.756 & 0.953 & 0.901 & 0.347 & 0.469 \\ \hline
\end{tabular}
}
\caption{Average pixel-wise and object-wise metrics for the \textit{intermittentObjectMotion} category of the CDNet dataset.}
\label{table:results_cdnet_intermittentobjectmotion}
\end{table*}

\begin{table*}[t]
\centering
\resizebox{1\textwidth}{!}{
\begin{tabular}{lccccccccccccccc}
\hline
Method                      & a contrario validation & & \multicolumn{5}{c}{pixel-wise} & \multicolumn{7}{c}{object-wise} \\

                            \cmidrule[0.4pt](lr{0.125em}){3-7}%
                            \cmidrule[0.4pt](lr{0.125em}){8-16}%
                            &                        & \re$^{pi}\uparrow$    & \fpr$^{pi}\downarrow$   & \pwc$^{pi}\downarrow$   & \pr$^{pi}\uparrow$    & \fe$^{pi}\uparrow$   & \re$^{ob}\uparrow$     & \fpr$^{ob}\downarrow$     & \pwc$^{ob}\downarrow$    & \pr$^{ob}\uparrow$     & \fe$^{ob}\uparrow$   & $sIoU\uparrow$ & \fe$^{sIoU}\uparrow$ \\

                            \cmidrule[0.4pt](lr{0.125em}){3-3}%
                            \cmidrule[0.4pt](lr{0.125em}){4-4}%
                            \cmidrule[0.4pt](lr{0.125em}){5-5}%
                            \cmidrule[0.4pt](lr{0.125em}){6-6}%
                            \cmidrule[0.4pt](lr{0.125em}){7-7}%
                            \cmidrule[0.4pt](lr{0.125em}){8-8}%
                            \cmidrule[0.4pt](lr{0.125em}){9-9}%
                            \cmidrule[0.4pt](lr{0.125em}){10-10}%
                            \cmidrule[0.4pt](lr{0.125em}){11-11}%
                            \cmidrule[0.4pt](lr{0.125em}){12-12}%
                            \cmidrule[0.4pt](lr{0.125em}){13-13}%
                            \cmidrule[0.4pt](lr{0.125em}){14-14}%
                            \cmidrule[0.4pt](lr{0.125em}){15-15}%
                            \cmidrule[0.4pt](lr{0.125em}){16-16}%
ViBe~\cite{vibe}                              & \cross    & 0.392 & 0.006 & 2.405 & 0.587 & 0.379 & 0.885 & 36.375 & 97.026 & 0.030 & 0.057 & 0.008 & 0.013 \\
& \checkmark   & 0.361 & 0.001 & 1.938 & 0.764 & 0.456 & 0.755 & 6.724 & 92.355 & 0.078 & 0.139 & 0.020 & 0.031 \\ \hline
SuBSENSE~\cite{subsense}                      & \cross       & 0.801 & 0.007 & 1.397 & 0.642 & 0.646 & 0.853 & 1.192 & 70.970 & 0.302 & 0.417 & 0.151 & 0.258 \\
& \checkmark   & 0.633 & 0.006 & 1.675 & 0.607 & 0.579 & 0.742 & 0.893 & 70.631 & 0.319 & 0.427 & 0.155 & 0.253 \\ \hline
SemanticBGS (with ViBe)~\cite{semanticbgs}      & \cross     & 0.428 & 0.006 & 2.193 & 0.617 & 0.427 & 0.916 & 38.068 & 97.504 & 0.025 & 0.048 & 0.008 & 0.015 \\
& \checkmark   & 0.377 & 0.001 & 1.829 & 0.772 & 0.471 & 0.755 & 6.408 & 91.837 & 0.084 & 0.148 & 0.023 & 0.039 \\ \hline
SemanticBGS (with SuBSENSE)~\cite{semanticbgs}  & \cross     & 0.802 & 0.005 & 1.142 & 0.675 & 0.672 & 0.856 & 1.502 & 75.903 & 0.248 & 0.370 & 0.121 & 0.240 \\
& \checkmark   & 0.634 & 0.003 & 1.430 & 0.647 & 0.597 & 0.744 & 0.911 & 70.832 & 0.317 & 0.426 & 0.150 & 0.256 \\ \hline
BSUV-Net~\cite{bsuv}                           & \cross      & 0.706 & 0.003 & 0.499 & 0.702 & 0.697 & 0.722 & 0.987 & 70.101 & 0.307 & 0.414 & 0.185 & 0.297 \\
& \checkmark   & 0.580 & 0.001 & 0.668 & 0.939 & 0.630 & 0.644 & 0.710 & 68.925 & 0.343 & 0.430 & 0.206 & 0.305 \\ \hline
BSUV-Net 2.0~\cite{bsuv2.0}                   & \cross       & 0.539 & 0.001 & 0.848 & 0.887 & 0.611 & 0.565 & 0.553 & 70.336 & 0.385 & 0.422 & 0.215 & 0.287 \\
& \checkmark   & 0.537 & 0.001 & 0.852 & 0.892 & 0.609 & 0.559 & 0.447 & 67.164 & 0.598 & 0.465 & 0.300 & 0.310 \\ \hline
BSUV-Net FPM~\cite{bsuv}                    & \cross         & 0.535 & 0.002 & 0.840 & 0.843 & 0.569 & 0.605 & 0.580 & 68.871 & 0.469 & 0.430 & 0.208 & 0.269 \\
& \checkmark   & 0.534 & 0.002 & 0.839 & 0.830 & 0.568 & 0.595 & 0.512 & 67.800 & 0.530 & 0.441 & 0.231 & 0.282 \\ \hline
\end{tabular}
}
\caption{Average pixel-wise and object-wise metrics for the \textit{lowFramerate} category of the CDNet dataset.}
\label{table:results_cdnet_lowframerate}
\end{table*}

\begin{table*}[t]
\centering
\resizebox{1\textwidth}{!}{
\begin{tabular}{lccccccccccccccc}
\hline
Method                      & a contrario validation & & \multicolumn{5}{c}{pixel-wise} & \multicolumn{7}{c}{object-wise} \\

                            \cmidrule[0.4pt](lr{0.125em}){3-7}%
                            \cmidrule[0.4pt](lr{0.125em}){8-16}%
                            &                        & \re$^{pi}\uparrow$    & \fpr$^{pi}\downarrow$   & \pwc$^{pi}\downarrow$   & \pr$^{pi}\uparrow$    & \fe$^{pi}\uparrow$   & \re$^{ob}\uparrow$     & \fpr$^{ob}\downarrow$     & \pwc$^{ob}\downarrow$    & \pr$^{ob}\uparrow$     & \fe$^{ob}\uparrow$   & $sIoU\uparrow$ & \fe$^{sIoU}\uparrow$ \\

                            \cmidrule[0.4pt](lr{0.125em}){3-3}%
                            \cmidrule[0.4pt](lr{0.125em}){4-4}%
                            \cmidrule[0.4pt](lr{0.125em}){5-5}%
                            \cmidrule[0.4pt](lr{0.125em}){6-6}%
                            \cmidrule[0.4pt](lr{0.125em}){7-7}%
                            \cmidrule[0.4pt](lr{0.125em}){8-8}%
                            \cmidrule[0.4pt](lr{0.125em}){9-9}%
                            \cmidrule[0.4pt](lr{0.125em}){10-10}%
                            \cmidrule[0.4pt](lr{0.125em}){11-11}%
                            \cmidrule[0.4pt](lr{0.125em}){12-12}%
                            \cmidrule[0.4pt](lr{0.125em}){13-13}%
                            \cmidrule[0.4pt](lr{0.125em}){14-14}%
                            \cmidrule[0.4pt](lr{0.125em}){15-15}%
                            \cmidrule[0.4pt](lr{0.125em}){16-16}%
ViBe~\cite{vibe}                              & \cross    & 0.365 & 0.008 & 2.169 & 0.481 & 0.388 & 0.919 & 26.915 & 97.482 & 0.025 & 0.049 & 0.006 & 0.006 \\
& \checkmark   & 0.290 & 0.006 & 2.028 & 0.574 & 0.338 & 0.668 & 3.790 & 89.097 & 0.151 & 0.193 & 0.027 & 0.024 \\ \hline
SuBSENSE~\cite{subsense}                      & \cross       & 0.625 & 0.018 & 2.625 & 0.439 & 0.490 & 0.873 & 1.804 & 76.394 & 0.243 & 0.376 & 0.062 & 0.077 \\
& \checkmark   & 0.477 & 0.011 & 2.131 & 0.506 & 0.446 & 0.643 & 0.880 & 76.328 & 0.316 & 0.368 & 0.097 & 0.092 \\ \hline
SemanticBGS (with ViBe)~\cite{semanticbgs}      & \cross     & 0.246 & 0.004 & 1.930 & 0.582 & 0.323 & 0.774 & 18.354 & 96.705 & 0.034 & 0.063 & 0.006 & 0.007 \\
& \checkmark   & 0.181 & 0.002 & 1.881 & 0.660 & 0.264 & 0.542 & 3.020 & 89.400 & 0.156 & 0.189 & 0.026 & 0.021 \\ \hline
SemanticBGS (with SuBSENSE)~\cite{semanticbgs}  & \cross     & 0.458 & 0.011 & 2.172 & 0.532 & 0.433 & 0.741 & 1.843 & 80.308 & 0.215 & 0.325 & 0.053 & 0.067 \\
& \checkmark   & 0.322 & 0.006 & 1.869 & 0.601 & 0.377 & 0.529 & 0.836 & 79.811 & 0.298 & 0.327 & 0.089 & 0.080 \\ \hline
BSUV-Net~\cite{bsuv}                           & \cross      & 0.631 & 0.006 & 1.404 & 0.736 & 0.627 & 0.693 & 1.076 & 73.490 & 0.297 & 0.410 & 0.112 & 0.165 \\
& \checkmark   & 0.571 & 0.005 & 1.459 & 0.749 & 0.584 & 0.576 & 0.608 & 72.465 & 0.418 & 0.422 & 0.181 & 0.190 \\ \hline
BSUV-Net 2.0~\cite{bsuv2.0}                   & \cross       & 0.464 & 0.001 & 1.208 & 0.934 & 0.558 & 0.394 & 0.601 & 80.007 & 0.359 & 0.325 & 0.163 & 0.199 \\
& \checkmark   & 0.437 & 0.001 & 1.263 & 0.936 & 0.535 & 0.337 & 0.375 & 79.779 & 0.444 & 0.324 & 0.234 & 0.218 \\ \hline
BSUV-Net FPM~\cite{bsuv}                    & \cross         & 0.517 & 0.002 & 1.252 & 0.860 & 0.607 & 0.529 & 0.674 & 73.162 & 0.391 & 0.421 & 0.173 & 0.222 \\
& \checkmark   & 0.450 & 0.001 & 1.338 & 0.878 & 0.545 & 0.437 & 0.454 & 74.976 & 0.456 & 0.391 & 0.219 & 0.219 \\ \hline
\end{tabular}
}
\caption{Average pixel-wise and object-wise metrics for the \textit{nightVideos} category of the CDNet dataset.}
\label{table:results_cdnet_nightvideos}
\end{table*}

\begin{table*}[t]
\centering
\resizebox{1\textwidth}{!}{
\begin{tabular}{lccccccccccccccc}
\hline
Method                      & a contrario validation & & \multicolumn{5}{c}{pixel-wise} & \multicolumn{7}{c}{object-wise} \\

                            \cmidrule[0.4pt](lr{0.125em}){3-7}%
                            \cmidrule[0.4pt](lr{0.125em}){8-16}%
                            &                        & \re$^{pi}\uparrow$    & \fpr$^{pi}\downarrow$   & \pwc$^{pi}\downarrow$   & \pr$^{pi}\uparrow$    & \fe$^{pi}\uparrow$   & \re$^{ob}\uparrow$     & \fpr$^{ob}\downarrow$     & \pwc$^{ob}\downarrow$    & \pr$^{ob}\uparrow$     & \fe$^{ob}\uparrow$   & $sIoU\uparrow$ & \fe$^{sIoU}\uparrow$ \\

                            \cmidrule[0.4pt](lr{0.125em}){3-3}%
                            \cmidrule[0.4pt](lr{0.125em}){4-4}%
                            \cmidrule[0.4pt](lr{0.125em}){5-5}%
                            \cmidrule[0.4pt](lr{0.125em}){6-6}%
                            \cmidrule[0.4pt](lr{0.125em}){7-7}%
                            \cmidrule[0.4pt](lr{0.125em}){8-8}%
                            \cmidrule[0.4pt](lr{0.125em}){9-9}%
                            \cmidrule[0.4pt](lr{0.125em}){10-10}%
                            \cmidrule[0.4pt](lr{0.125em}){11-11}%
                            \cmidrule[0.4pt](lr{0.125em}){12-12}%
                            \cmidrule[0.4pt](lr{0.125em}){13-13}%
                            \cmidrule[0.4pt](lr{0.125em}){14-14}%
                            \cmidrule[0.4pt](lr{0.125em}){15-15}%
                            \cmidrule[0.4pt](lr{0.125em}){16-16}%
ViBe~\cite{vibe}                              & \cross    & 0.368 & 0.001 & 4.259 & 0.978 & 0.515 & 0.722 & 12.493 & 90.416 & 0.103 & 0.168 & 0.019 & 0.059 \\
& \checkmark   & 0.360 & 0.000 & 4.283 & 0.982 & 0.506 & 0.705 & 1.361 & 68.358 & 0.389 & 0.475 & 0.042 & 0.122 \\ \hline
SuBSENSE~\cite{subsense}                      & \cross       & 0.740 & 0.009 & 1.907 & 0.869 & 0.783 & 0.755 & 0.836 & 49.920 & 0.608 & 0.646 & 0.221 & 0.323 \\
& \checkmark   & 0.739 & 0.008 & 1.881 & 0.871 & 0.783 & 0.754 & 0.650 & 46.719 & 0.652 & 0.678 & 0.243 & 0.345 \\ \hline
SemanticBGS (with ViBe)~\cite{semanticbgs}      & \cross     & 0.469 & 0.004 & 2.789 & 0.968 & 0.607 & 0.721 & 7.818 & 88.086 & 0.129 & 0.208 & 0.032 & 0.115 \\
& \checkmark   & 0.462 & 0.004 & 2.808 & 0.970 & 0.600 & 0.709 & 0.768 & 59.856 & 0.524 & 0.572 & 0.082 & 0.242 \\ \hline
SemanticBGS (with SuBSENSE)~\cite{semanticbgs}  & \cross     & 0.694 & 0.009 & 1.904 & 0.891 & 0.764 & 0.730 & 0.697 & 49.361 & 0.665 & 0.655 & 0.231 & 0.336 \\
& \checkmark   & 0.693 & 0.009 & 1.886 & 0.892 & 0.764 & 0.728 & 0.366 & 42.011 & 0.765 & 0.723 & 0.282 & 0.386 \\ \hline
BSUV-Net~\cite{bsuv}                           & \cross      & 0.907 & 0.004 & 0.993 & 0.913 & 0.906 & 0.783 & 0.510 & 41.519 & 0.705 & 0.724 & 0.412 & 0.530 \\
& \checkmark   & 0.907 & 0.004 & 0.989 & 0.914 & 0.907 & 0.780 & 0.231 & 33.768 & 0.829 & 0.787 & 0.493 & 0.565 \\ \hline
BSUV-Net 2.0~\cite{bsuv2.0}                   & \cross       & 0.568 & 0.001 & 1.376 & 0.950 & 0.657 & 0.498 & 0.214 & 59.187 & 0.693 & 0.531 & 0.332 & 0.416 \\
& \checkmark   & 0.567 & 0.001 & 1.378 & 0.950 & 0.656 & 0.495 & 0.014 & 51.269 & 0.964 & 0.598 & 0.420 & 0.440 \\ \hline
BSUV-Net FPM~\cite{bsuv}                    & \cross         & 0.411 & 0.002 & 2.327 & 0.917 & 0.507 & 0.471 & 0.089 & 58.129 & 0.848 & 0.560 & 0.323 & 0.339 \\
& \checkmark   & 0.410 & 0.002 & 2.327 & 0.918 & 0.506 & 0.462 & 0.018 & 55.032 & 0.960 & 0.581 & 0.373 & 0.352 \\ \hline
\end{tabular}
}
\caption{Average pixel-wise and object-wise metrics for the \textit{thermal} category of the CDNet dataset.}
\label{table:results_cdnet_thermal}
\end{table*}

\begin{table*}[t]
\centering
\resizebox{1\textwidth}{!}{
\begin{tabular}{lccccccccccccccc}
\hline
Method                      & a contrario validation & & \multicolumn{5}{c}{pixel-wise} & \multicolumn{7}{c}{object-wise} \\

                            \cmidrule[0.4pt](lr{0.125em}){3-7}%
                            \cmidrule[0.4pt](lr{0.125em}){8-16}%
                            &                        & \re$^{pi}\uparrow$    & \fpr$^{pi}\downarrow$   & \pwc$^{pi}\downarrow$   & \pr$^{pi}\uparrow$    & \fe$^{pi}\uparrow$   & \re$^{ob}\uparrow$     & \fpr$^{ob}\downarrow$     & \pwc$^{ob}\downarrow$    & \pr$^{ob}\uparrow$     & \fe$^{ob}\uparrow$   & $sIoU\uparrow$ & \fe$^{sIoU}\uparrow$ \\

                            \cmidrule[0.4pt](lr{0.125em}){3-3}%
                            \cmidrule[0.4pt](lr{0.125em}){4-4}%
                            \cmidrule[0.4pt](lr{0.125em}){5-5}%
                            \cmidrule[0.4pt](lr{0.125em}){6-6}%
                            \cmidrule[0.4pt](lr{0.125em}){7-7}%
                            \cmidrule[0.4pt](lr{0.125em}){8-8}%
                            \cmidrule[0.4pt](lr{0.125em}){9-9}%
                            \cmidrule[0.4pt](lr{0.125em}){10-10}%
                            \cmidrule[0.4pt](lr{0.125em}){11-11}%
                            \cmidrule[0.4pt](lr{0.125em}){12-12}%
                            \cmidrule[0.4pt](lr{0.125em}){13-13}%
                            \cmidrule[0.4pt](lr{0.125em}){14-14}%
                            \cmidrule[0.4pt](lr{0.125em}){15-15}%
                            \cmidrule[0.4pt](lr{0.125em}){16-16}%
ViBe~\cite{vibe}                              & \cross    & 0.652 & 0.001 & 1.571 & 0.960 & 0.773 & 0.951 & 15.416 & 93.534 & 0.065 & 0.119 & 0.022 & 0.079 \\
& \checkmark   & 0.646 & 0.001 & 1.572 & 0.975 & 0.773 & 0.941 & 2.511 & 67.621 & 0.328 & 0.458 & 0.060 & 0.178 \\ \hline
SuBSENSE~\cite{subsense}                      & \cross       & 0.903 & 0.008 & 1.233 & 0.857 & 0.879 & 0.913 & 0.282 & 35.656 & 0.683 & 0.773 & 0.368 & 0.519 \\
& \checkmark   & 0.902 & 0.006 & 1.081 & 0.877 & 0.889 & 0.909 & 0.153 & 26.006 & 0.794 & 0.846 & 0.416 & 0.562 \\ \hline
SemanticBGS (with ViBe)~\cite{semanticbgs}      & \cross     & 0.909 & 0.000 & 0.421 & 0.988 & 0.946 & 0.947 & 6.041 & 80.790 & 0.194 & 0.303 & 0.089 & 0.202 \\
& \checkmark   & 0.905 & 0.000 & 0.433 & 0.990 & 0.945 & 0.935 & 1.491 & 57.201 & 0.444 & 0.572 & 0.199 & 0.394 \\ \hline
SemanticBGS (with SuBSENSE)~\cite{semanticbgs}  & \cross     & 0.967 & 0.003 & 0.379 & 0.948 & 0.958 & 0.922 & 0.317 & 35.489 & 0.685 & 0.775 & 0.419 & 0.584 \\
& \checkmark   & 0.965 & 0.002 & 0.381 & 0.950 & 0.957 & 0.917 & 0.147 & 23.100 & 0.828 & 0.867 & 0.512 & 0.662 \\ \hline
BSUV-Net~\cite{bsuv}                           & \cross      & 0.954 & 0.002 & 0.323 & 0.954 & 0.954 & 0.922 & 0.293 & 30.888 & 0.729 & 0.803 & 0.463 & 0.639 \\
& \checkmark   & 0.952 & 0.002 & 0.328 & 0.955 & 0.953 & 0.914 & 0.146 & 22.167 & 0.831 & 0.864 & 0.535 & 0.689 \\ \hline
BSUV-Net 2.0~\cite{bsuv2.0}                   & \cross       & 0.933 & 0.002 & 0.497 & 0.968 & 0.950 & 0.862 & 0.307 & 36.776 & 0.737 & 0.768 & 0.492 & 0.683 \\
& \checkmark   & 0.930 & 0.002 & 0.516 & 0.968 & 0.948 & 0.854 & 0.099 & 24.055 & 0.890 & 0.863 & 0.585 & 0.743 \\ \hline
BSUV-Net FPM~\cite{bsuv}                    & \cross         & 0.953 & 0.001 & 0.340 & 0.973 & 0.963 & 0.888 & 0.102 & 22.995 & 0.862 & 0.867 & 0.571 & 0.725 \\
& \checkmark   & 0.952 & 0.001 & 0.345 & 0.973 & 0.962 & 0.883 & 0.051 & 17.967 & 0.926 & 0.899 & 0.617 & 0.751 \\ \hline
\end{tabular}
}
\caption{Average pixel-wise and object-wise metrics for the \textit{shadow} category of the CDNet dataset.}
\label{table:results_cdnet_shadow}
\end{table*}

\begin{table*}[t]
\centering
\resizebox{1\textwidth}{!}{
\begin{tabular}{lccccccccccccccc}
\hline
Method                      & a contrario validation & & \multicolumn{5}{c}{pixel-wise} & \multicolumn{7}{c}{object-wise} \\

                            \cmidrule[0.4pt](lr{0.125em}){3-7}%
                            \cmidrule[0.4pt](lr{0.125em}){8-16}%
                            &                        & \re$^{pi}\uparrow$    & \fpr$^{pi}\downarrow$   & \pwc$^{pi}\downarrow$   & \pr$^{pi}\uparrow$    & \fe$^{pi}\uparrow$   & \re$^{ob}\uparrow$     & \fpr$^{ob}\downarrow$     & \pwc$^{ob}\downarrow$    & \pr$^{ob}\uparrow$     & \fe$^{ob}\uparrow$   & $sIoU\uparrow$ & \fe$^{sIoU}\uparrow$ \\

                            \cmidrule[0.4pt](lr{0.125em}){3-3}%
                            \cmidrule[0.4pt](lr{0.125em}){4-4}%
                            \cmidrule[0.4pt](lr{0.125em}){5-5}%
                            \cmidrule[0.4pt](lr{0.125em}){6-6}%
                            \cmidrule[0.4pt](lr{0.125em}){7-7}%
                            \cmidrule[0.4pt](lr{0.125em}){8-8}%
                            \cmidrule[0.4pt](lr{0.125em}){9-9}%
                            \cmidrule[0.4pt](lr{0.125em}){10-10}%
                            \cmidrule[0.4pt](lr{0.125em}){11-11}%
                            \cmidrule[0.4pt](lr{0.125em}){12-12}%
                            \cmidrule[0.4pt](lr{0.125em}){13-13}%
                            \cmidrule[0.4pt](lr{0.125em}){14-14}%
                            \cmidrule[0.4pt](lr{0.125em}){15-15}%
                            \cmidrule[0.4pt](lr{0.125em}){16-16}%
ViBe~\cite{vibe}                              & \cross    & 0.698 & 0.001 & 0.286 & 0.720 & 0.691 & 0.981 & 22.312 & 96.774 & 0.032 & 0.062 & 0.018 & 0.038 \\
 & \checkmark   & 0.660 & 0.000 & 0.219 & 0.942 & 0.767 & 0.843 & 1.314 & 75.241 & 0.301 & 0.386 & 0.155 & 0.238 \\ \hline
SuBSENSE~\cite{subsense}                      & \cross       & 0.756 & 0.000 & 0.165 & 0.863 & 0.785 & 0.779 & 0.577 & 69.219 & 0.391 & 0.464 & 0.238 & 0.350 \\
 & \checkmark   & 0.733 & 0.000 & 0.154 & 0.900 & 0.793 & 0.729 & 0.410 & 68.278 & 0.466 & 0.477 & 0.298 & 0.374 \\ \hline
SemanticBGS (with ViBe)~\cite{semanticbgs}      & \cross     & 0.687 & 0.001 & 0.274 & 0.757 & 0.710 & 0.955 & 24.393 & 96.991 & 0.030 & 0.058 & 0.018 & 0.037 \\
 & \checkmark   & 0.651 & 0.000 & 0.226 & 0.957 & 0.766 & 0.827 & 1.192 & 75.164 & 0.309 & 0.389 & 0.158 & 0.238 \\ \hline
SemanticBGS (with SuBSENSE)~\cite{semanticbgs}  & \cross     & 0.967 & 0.749 & 0.000 & 0.160 & 0.899 & 0.806 & 0.774 & 8.405 & 76.720 & 0.240 & 0.352 & 0.150 & 0.270 \\
 & \checkmark   & 0.726 & 0.000 & 0.150 & 0.958 & 0.817 & 0.725 & 0.400 & 68.278 & 0.458 & 0.477 & 0.291 & 0.372 \\ \hline
BSUV-Net~\cite{bsuv}                           & \cross      & 0.735 & 0.002 & 0.346 & 0.753 & 0.665 & 0.838 & 0.805 & 67.266 & 0.395 & 0.479 & 0.224 & 0.332 \\
 & \checkmark   & 0.713 & 0.002 & 0.340 & 0.756 & 0.652 & 0.777 & 0.552 & 66.654 & 0.449 & 0.492 & 0.266 & 0.357 \\ \hline
BSUV-Net 2.0~\cite{bsuv2.0}                   & \cross       & 0.588 & 0.000 & 0.248 & 0.959 & 0.720 & 0.700 & 0.282 & 61.968 & 0.507 & 0.543 & 0.270 & 0.371 \\
 & \checkmark   & 0.566 & 0.000 & 0.250 & 0.961 & 0.703 & 0.640 & 0.158 & 59.825 & 0.620 & 0.566 & 0.344 & 0.398 \\ \hline
BSUV-Net FPM~\cite{bsuv}                    & \cross         & 0.527 & 0.001 & 0.297 & 0.766 & 0.550 & 0.626 & 0.278 & 67.232 & 0.516 & 0.483 & 0.244 & 0.299 \\
 & \checkmark   & 0.512 & 0.001 & 0.296 & 0.766 & 0.539 & 0.583 & 0.174 & 64.682 & 0.596 & 0.511 & 0.293 & 0.329 \\ \hline
\end{tabular}
}
\caption{Average pixel-wise and object-wise metrics for the \textit{turbulence} category of the CDNet dataset.}
\label{table:results_cdnet_turbulence}
\end{table*}

\subsection{LASIESTA dataset}
The full quantitative results for both pixel and object metrics are provided in Table~\ref{table:lasiesta_overall_results}.
\subsection{Sequences from from Zhong and S. Sclaroff}
Table~\ref{table:iccv03_results} shows the full quantitative results for both pixel and object metrics, for the sequences from Zhong and S. Sclaroff.

\begin{table*}[t]
\centering
\resizebox{1\textwidth}{!}{
\begin{tabular}{lccccccccccccccc}
Method                      & a contrario validation & & \multicolumn{5}{c}{pixel-wise} & \multicolumn{7}{c}{object-wise} \\

                            \cmidrule[0.4pt](lr{0.125em}){3-7}%
                            \cmidrule[0.4pt](lr{0.125em}){8-16}%
                            &                        & \re$^{pi}\uparrow$    & \fpr$^{pi}\downarrow$   & \pwc$^{pi}\downarrow$   & \pr$^{pi}\uparrow$    & \fe$^{pi}\uparrow$   & \re$^{ob}\uparrow$     & \fpr$^{ob}\downarrow$     & \pwc$^{ob}\downarrow$    & \pr$^{ob}\uparrow$     & \fe$^{ob}\uparrow$   & $sIoU\uparrow$ & \fe$^{sIoU}\uparrow$ \\

                            \cmidrule[0.4pt](lr{0.125em}){3-3}%
                            \cmidrule[0.4pt](lr{0.125em}){4-4}%
                            \cmidrule[0.4pt](lr{0.125em}){5-5}%
                            \cmidrule[0.4pt](lr{0.125em}){6-6}%
                            \cmidrule[0.4pt](lr{0.125em}){7-7}%
                            \cmidrule[0.4pt](lr{0.125em}){8-8}%
                            \cmidrule[0.4pt](lr{0.125em}){9-9}%
                            \cmidrule[0.4pt](lr{0.125em}){10-10}%
                            \cmidrule[0.4pt](lr{0.125em}){11-11}%
                            \cmidrule[0.4pt](lr{0.125em}){12-12}%
                            \cmidrule[0.4pt](lr{0.125em}){13-13}%
                            \cmidrule[0.4pt](lr{0.125em}){14-14}%
                            \cmidrule[0.4pt](lr{0.125em}){15-15}%
                            \cmidrule[0.4pt](lr{0.125em}){16-16}%
ViBe~\cite{vibe} & \cross            & \textbf{0.659} & 0.009 & 1.656 & 0.714 & 0.655 & \textbf{0.893} & 11.320 & 77.503 & 0.178 & 0.270 & 0.0757	& 0.2256 \\
             & \checkmark            & 0.656 & 0.009 & \textbf{1.643} & \textbf{0.721} & \textbf{0.657} & 0.889 & \textbf{4.163}  & \textbf{51.370} & \textbf{0.451} & \textbf{0.552} & \textbf{0.1202} &	\textbf{0.3266}\\  \hline
SuBSENSE~\cite{subsense} & \cross    & 0.826 & 0.012 & \textbf{1.360} & 0.762 & 0.763 & \textbf{0.799} & 0.231  & 28.136 & 0.779 & 0.767 & 0.4761	& 0.6284 \\
 & \checkmark                        & 0.826 & 0.012 & 1.361 & 0.762 & 0.763 & 0.768 & \textbf{0.213}  & \textbf{27.795} & \textbf{0.786} & \textbf{0.770} & \textbf{0.4809}	 & \textbf{0.6307} \\ \hline
BSUV-Net 2.0~\cite{bsuv2.0} & \cross & 0.823 & 0.001 & 0.377 & 0.901 & 0.853 & \textbf{0.701} & 0.110  & 35.241 & 0.790 & 0.725  & 0.5996	 & 0.7716\\
             & \checkmark            & 0.823 & 0.001 & 0.377 & 0.901 & 0.853 & 0.699 & \textbf{0.054}  & \textbf{30.580} & \textbf{0.866} & \textbf{0.758}  & \textbf{0.6455}	 & \textbf{0.7925}\\ \hline
BSUV-Net FPM~\cite{bsuv} & \cross    & 0.840 & 0.002 & 0.378 & 0.885 & 0.854 & \textbf{0.735} & 0.079  & 28.864 & 0.837 & 0.769  & 0.6124	 & 0.7608\\
             & \checkmark            & 0.840 & 0.002 & 0.378 & 0.885 & \textbf{0.885} & 0.733 & \textbf{0.041}  & \textbf{26.250} & \textbf{0.884} & \textbf{0.787} & \textbf{0.6448}	 & \textbf{0.7747} \\ \hline
BSUV-Net~\cite{bsuv}     & \cross    & 0.871 & 0.002 & 0.301 & 0.870 & 0.865 & 0.820 & 0.137  & 25.275 & 0.796 & 0.796  & 0.5896	 & 0.7470\\
             & \checkmark            & 0.871 & 0.002 & 0.301 & 0.870 & 0.865 & 0.820 & \textbf{0.067}  & \textbf{20.436} & \textbf{0.853} & \textbf{0.829}  & \textbf{0.6494}	 & \textbf{0.7722}\\ \hline
\end{tabular}
}

\caption{Average pixel-wise and object-wise metrics for all evaluated sequences in LASIESTA dataset.}
\label{table:lasiesta_overall_results}
\end{table*}

\begin{table*}[t]
\centering
\resizebox{1\textwidth}{!}{
\begin{tabular}{lccccccccccccccc}
\hline
Method                      & a contrario validation & & \multicolumn{5}{c}{pixel-wise} & \multicolumn{7}{c}{object-wise} \\

                            \cmidrule[0.4pt](lr{0.125em}){3-7}%
                            \cmidrule[0.4pt](lr{0.125em}){8-16}%
                            &                        & \re$^{pi}\uparrow$    & \fpr$^{pi}\downarrow$   & \pwc$^{pi}\downarrow$   & \pr$^{pi}\uparrow$    & \fe$^{pi}\uparrow$   & \re$^{ob}\uparrow$     & \fpr$^{ob}\downarrow$     & \pwc$^{ob}\downarrow$    & \pr$^{ob}\uparrow$     & \fe$^{ob}\uparrow$   & $sIoU\uparrow$ & \fe$^{sIoU}\uparrow$ \\

                            \cmidrule[0.4pt](lr{0.125em}){3-3}%
                            \cmidrule[0.4pt](lr{0.125em}){4-4}%
                            \cmidrule[0.4pt](lr{0.125em}){5-5}%
                            \cmidrule[0.4pt](lr{0.125em}){6-6}%
                            \cmidrule[0.4pt](lr{0.125em}){7-7}%
                            \cmidrule[0.4pt](lr{0.125em}){8-8}%
                            \cmidrule[0.4pt](lr{0.125em}){9-9}%
                            \cmidrule[0.4pt](lr{0.125em}){10-10}%
                            \cmidrule[0.4pt](lr{0.125em}){11-11}%
                            \cmidrule[0.4pt](lr{0.125em}){12-12}%
                            \cmidrule[0.4pt](lr{0.125em}){13-13}%
                            \cmidrule[0.4pt](lr{0.125em}){14-14}%
                            \cmidrule[0.4pt](lr{0.125em}){15-15}%
                            \cmidrule[0.4pt](lr{0.125em}){16-16}%
ViBe~\cite{vibe} & \cross &             \textbf{0.401} & 0.048 & 5.461 & 0.192 & 0.240 & 0.543  & 74.323 & 99.093 & 0.009  & 0.02   & 0.004 & 	0.004 \\
             & \checkmark &             0.347 & \textbf{0.002} & \textbf{0.913} & \textbf{0.743} & \textbf{0.440} & \textbf{0.860}  & \textbf{0.156}  & \textbf{36.152} & \textbf{0.681}  & \textbf{0.75}  & \textbf{0.088}	 & \textbf{0.134}  \\  \hline
SuBSENSE~\cite{subsense}     & \cross & \textbf{0.488} & 0.024 & 2.911 & 0.289 & 0.359 & \textbf{0.890}  & 0.728  & 67.240 & 0.352  & 0.49   & 0.059	 & 0.089 \\
             & \checkmark &             0.483 & \textbf{0.003} & \textbf{0.871} & \textbf{0.778} & \textbf{0.469} & 0.830  & \textbf{0.035}  & \textbf{23.173} & \textbf{0.878}  & \textbf{0.85}  & \textbf{0.285}	 & \textbf{0.272}  \\  \hline
BSUV-Net 2.0~\cite{bsuv2.0} & \cross &  \textbf{0.438} & 0.001 & \textbf{0.693} & 0.883 & \textbf{0.569} & \textbf{0.655}  & 0.203  & 52.954 & 0.604  & 0.612  v0.295 & 	0.442\\
             & \checkmark &             0.411 & \textbf{0.0002} & 0.695 & \textbf{0.960} & 0.541 & 0.585  & \textbf{0.000}  & \textbf{41.500} & \textbf{1.000}  & \textbf{0.724}   & \textbf{0.536}	 & \textbf{0.577}\\  \hline
BSUV-Net FPM~\cite{bsuv} & \cross &     \textbf{0.847} & 0.030 & 3.115 & 0.513 & 0.545 & \textbf{0.918}  & 0.228  & 38.999 & 0.658  & 0.738  & 0.331 & 	0.509 \\
             & \checkmark &             0.787 & \textbf{0.004} & \textbf{0.593} & \textbf{0.742} & \textbf{0.758} & 0.787  & \textbf{0.000}  & \textbf{21.292} & \textbf{1.000}  & \textbf{0.879} & \textbf{0.662}	 & \textbf{0.717}  \\  \hline
BSUV-Net~\cite{bsuv}     & \cross &     \textbf{0.908} & 0.048 & 4.877 & 0.470 & 0.528 & \textbf{0.929}  & 1.290  & 49.913 & 0.520  & 0.565  & 0.274	 & 0.420 \\
             & \checkmark &             0.855 & \textbf{0.009} & \textbf{1.075} & \textbf{0.601} & \textbf{0.686} & 0.839  & \textbf{0.006}  & \textbf{17.548} & \textbf{0.974}  & \textbf{0.900}  & \textbf{0.467}	 & \textbf{0.484} \\  \hline
\end{tabular}
}
\caption{Average pixel-wise and object-wise metrics for the sequences in Zhong and S. Sclaroff~\cite{iccv03_paper}, namely \textit{water} and \textit{escalator}.}
\label{table:results_iccv}
\end{table*}


\section{Complementary examples on the separability of FP and TP by NFA score}
\label{supmat_C}
\begin{figure}[b]
    \centering
    \includegraphics[width=0.8\linewidth]{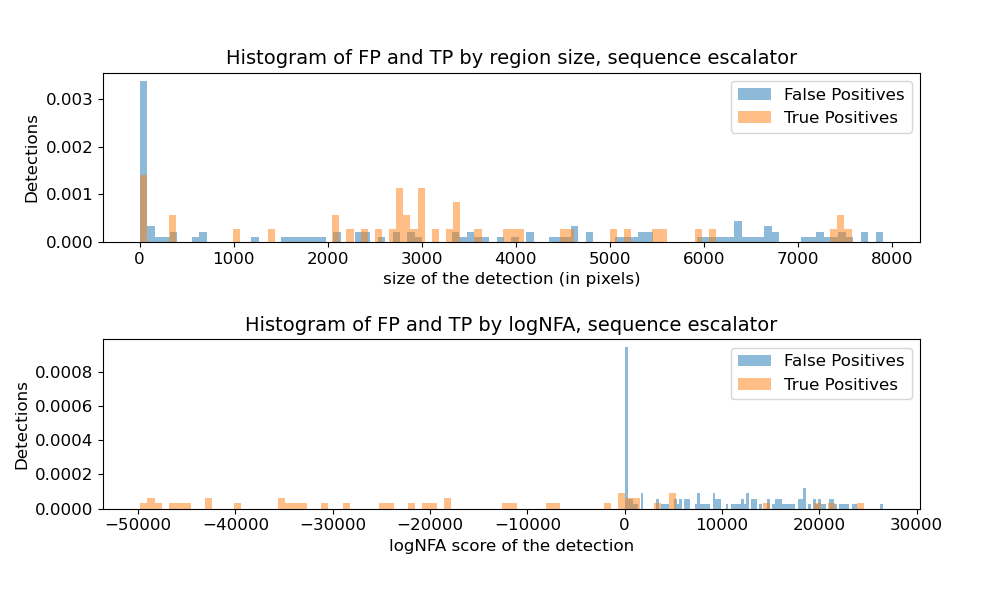}
    \vspace{-2em}
    \caption{Comparison of the histograms of TP and FP computed by the BSUVFPM algorithm for the sequence \textit{escalator}, by size of the detection (top) and by the $\log(\mathrm{NFA})$ score (bottom).}
    \label{fig:histograms_ex1}
\end{figure}

\begin{figure}[b]
    \centering
    \includegraphics[width=0.8\linewidth]{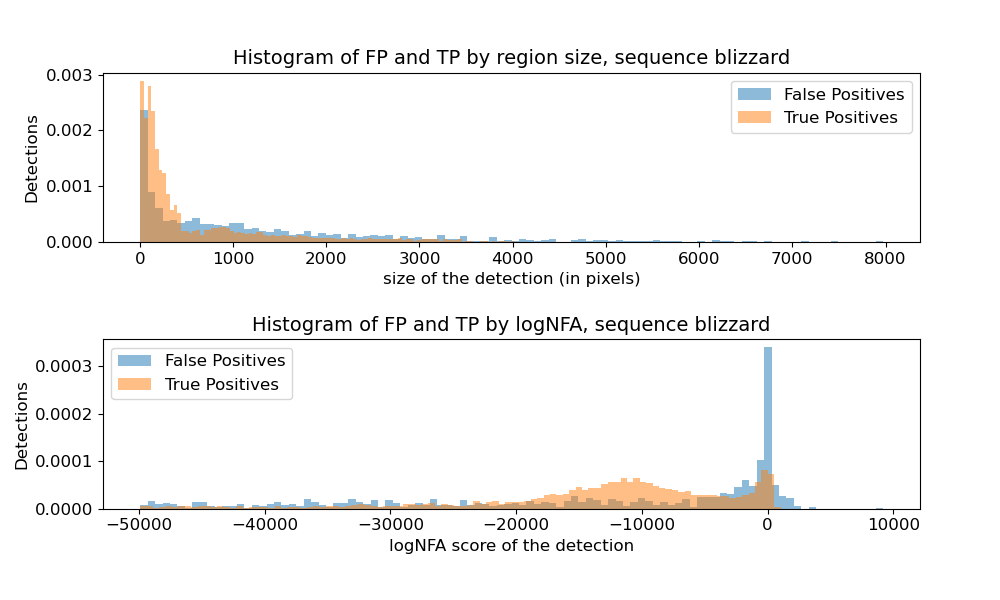}
    \vspace{-2em}
    \caption{Comparison of the histograms of TP and FP computed by the SemanticBGS algorithm (with SuBSENSE) for the sequence \textit{blizzard}, by size of the detection (top) and by the $\log(\mathrm{NFA})$ score (bottom).}
    \label{fig:histograms_ex2}
\end{figure}

\begin{figure}[t]
    \centering
    \includegraphics[width=0.8\linewidth]{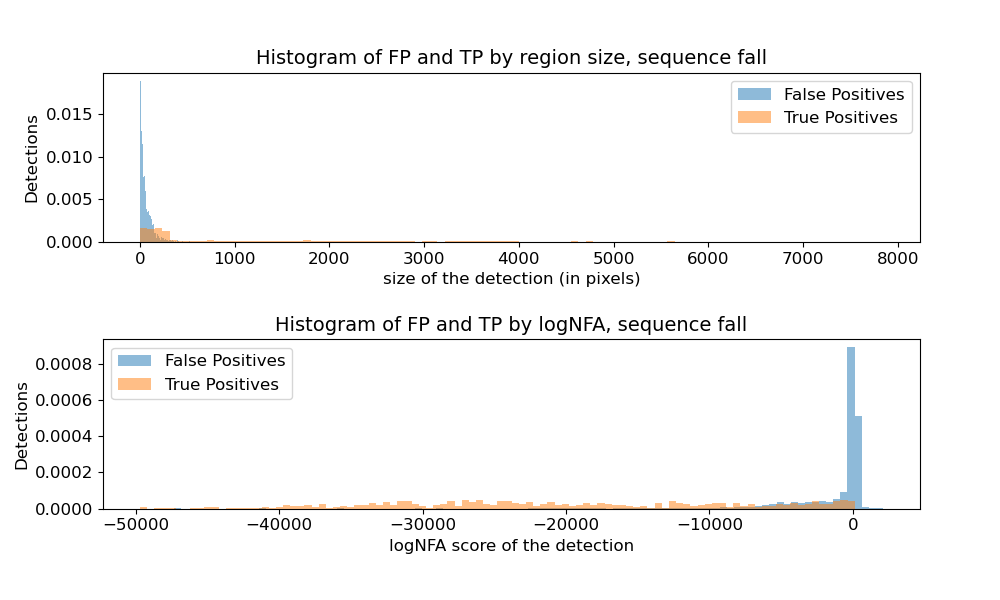}
    \vspace{-2em}
    \caption{Comparison of the histograms of TP and FP computed by the SuBSENSE algorithm for the sequence \textit{fall}, by size of the detection (top) and by the $\log(\mathrm{NFA})$ score (bottom).}
    \label{fig:histograms_ex3}
\end{figure}

\begin{figure}[t]
    \centering
    \includegraphics[width=0.8\linewidth]{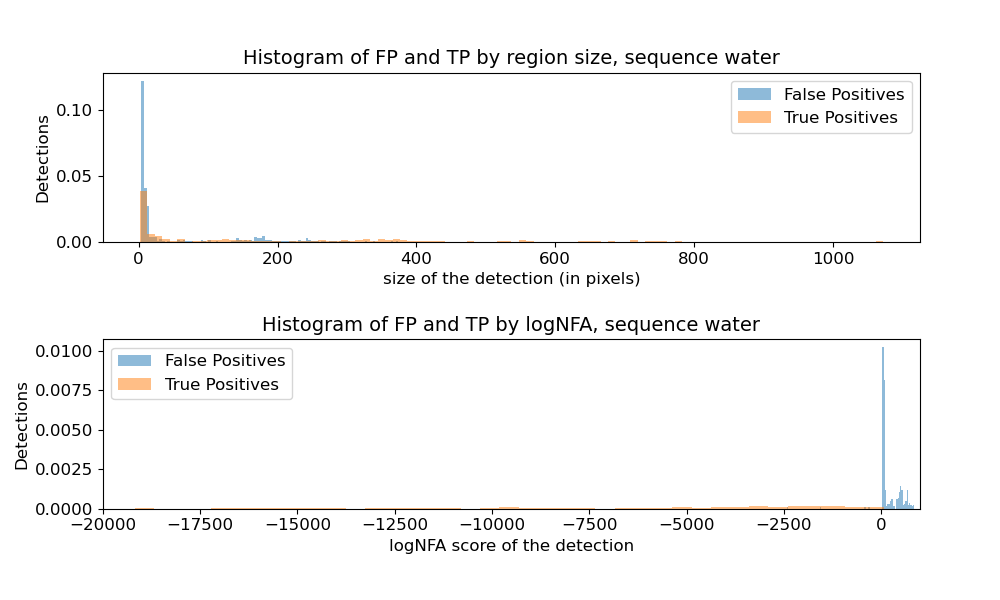}
    \caption{Comparison of the histograms of TP and FP computed by the ViBe algorithm for the sequence \textit{water}, by size of the detection (top) and by the $\log(\mathrm{NFA})$ score (bottom).}
    \label{fig:histograms_ex4}
\end{figure}

During the analysis of the impact of the a-contrario validation, we evaluate the separability of true and false positives by the computed $logNFA$ score. Moreover, we compare it with the naive approach of filtering small detections and visualize both the region size and $logNFA$ score histograms of FP and TP. The shown example in Figure~\ref{fig:histograms} corresponds to the results for SuBSENSE on the \textit{escalator} sequence. In order to support our analysis, we provide additional examples of sequences where the separability by merely filtering out small detections is not efficient, while the $logNFA$ provides a more suitable metric. Figure~\ref{fig:histograms_ex1} shows the example for the BSUVFPM algorithm and the sequence \textit{escalator}. Figure~\ref{fig:histograms_ex2} corresponds to the results of the SemanticBGS algorithm with SuBSENSE for the sequence \textit{blizzard}. Figure~\ref{fig:histograms_ex3} displays the example for the SuBSENSE algorithm and the sequence \textit{fall}, and lastly Figure~\ref{fig:histograms_ex4} shows the case for the ViBe method and the sequence \textit{water}.

\end{document}